\documentclass[sigconf,natbib=false]{acmart}
\AtBeginDocument{%
  }

\copyrightyear{2026}
\acmYear{2026}
\setcopyright{cc}
\setcctype{by}
\acmConference[SAC '26]{The 41st ACM/SIGAPP Symposium on Applied Computing}{March 23--27, 2026}{Thessaloniki, Greece}
\acmBooktitle{The 41st ACM/SIGAPP Symposium on Applied Computing (SAC '26), March 23--27, 2026, Thessaloniki, Greece}
\acmPrice{}
\acmDOI{10.1145/3748522.3779821}
\acmISBN{979-8-4007-2294-3/2026/03}

\RequirePackage[
  datamodel=acmdatamodel,
  style=acmnumeric,
  ]{biblatex}

\usepackage{algorithm}
\usepackage{algorithmic}
\usepackage{bm}
\usepackage{colortbl}

\usepackage{subfigure}
\usepackage{xspace}
\usepackage{multirow}
\usepackage{makecell}

\usepackage{pifont}

\newcommand{\cmark}{\ding{51}}
\newcommand{\xmark}{\ding{55}}

\newcommand{\smalltitle}[1]{ {\noindent\textbf{#1.}\hspace{0mm}}}

\begin{document}


\title{Balanced Online Class-Incremental Learning via Dual Classifiers}

\renewcommand{\shorttitle}{BISON}








\author{Shunjie Wen}
\affiliation{%
  \institution{Inha University}
  \city{Incheon}
  \country{Republic of Korea}}
\email{wenshunjie@inha.edu}

\author{Thomas Heinis}
\affiliation{%
  \institution{Imperial College London}
  \city{London}
  \country{UK}}
\email{t.heinis@imperial.ac.uk}

\author{Dong-Wan Choi}
\authornote{Corresponding author.}
\affiliation{%
  \institution{Inha University}
  \city{Incheon}
  \country{Republic of Korea}}
\email{dchoi@inha.ac.kr}


\renewcommand{\shortauthors}{S. Wen et al.}

\begin{abstract}
Online class-incremental learning (OCIL) focuses on gradually learning new classes (called \textit{plasticity}) from a stream of data in a single-pass, while concurrently preserving knowledge of previously learned classes (called \textit{stability}). The primary challenge in OCIL lies in maintaining a good balance between the knowledge of old and new classes within the continually updated model. Most existing methods rely on \textit{explicit} knowledge interaction through \textit{experience replay}, and often employ \textit{exclusive} training separation to address bias problems. Nevertheless, it still remains a big challenge to achieve a well-balanced learner, as these methods often exhibit either reduced plasticity or limited stability due to difficulties in continually integrating knowledge in the OCIL setting. In this paper, we propose a novel replay-based method, called \textbf{B}alanced \textbf{I}nclusive \textbf{S}eparation for \textbf{O}nline i\textbf{N}cremental learning (BISON), which can achieve both high plasticity and stability, thus ensuring more balanced performance in OCIL. Our BISON method proposes an \textit{inclusive} training separation strategy using dual classifiers so that knowledge from both old and new classes can effectively be integrated into the model, while introducing \textit{implicit} approaches for transferring knowledge across the two classifiers. Extensive experimental evaluations over three widely-used OCIL benchmark datasets demonstrate the superiority of BISON, showing more balanced yet better performance compared to state-of-the-art replay-based OCIL methods.
\end{abstract}




\maketitle

\section{Introduction} \label{sec:intro}
Online class-incremental learning (OCIL) has attracted considerable attention in the deep learning community, enabling knowledge accumulation of new classes over time. Unlike its offline counterpart, which assumes access to a complete training set for each task, OCIL is given a single-pass stream where the learner is allowed to view each mini-batch of the task only once. This partial accessibility of data makes OCIL more appealing in practice but also brings a greater challenge in addressing the core problem of continual learning: \textit{stability-plasticity} dilemma \cite{dilemma/mermillod2013stability}, i.e., balancing knowledge preservation (stability) with knowledge acquisition (plasticity). Due to the limited training opportunities with on-the-fly samples in OCIL, the model can easily be biased toward new classes without a careful strategy of knowledge preservation. Conversely, a strong policy to preserve past information can severely impair the learning performance due to the small number of incoming samples.

\begin{figure}[t]
\centering
\includegraphics[width=0.8\columnwidth]{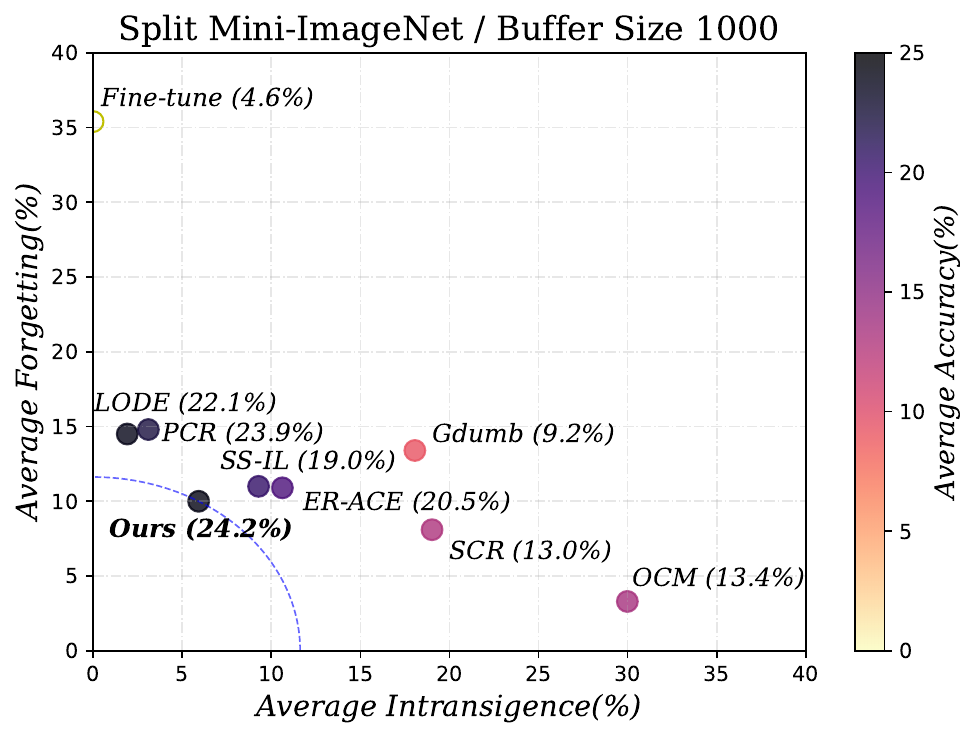}
\caption{Balance comparison in various OCIL methods on Split Mini-ImageNet ($\mathcal{M}$ = 1K). The color of the circle indicates the average accuracy. Lower forgetting and intransigence indicates better stability and higher plasticity, respectively. Our method (BISON) is closest to the bottom-left corner with the highest accuracy, implying the most balanced and outperforming performance.}
\label{fig:intro_balance}
\vspace{-3mm}
\end{figure}

The majority of OCIL methods tackle this challenge by using \textit{experience replay} (ER) \cite{ER/chaudhry2019tiny}, which trains a mix batch of incoming samples from the stream and previous ones that are partially stored in a memory buffer. However, this \textit{explicit} knowledge interaction via sample blending cannot completely resolve the stability-plasticity dilemma, since class imbalance is inevitable in OCIL with a fixed buffer size. As learning progresses over time, it would become more difficult to maintain balanced performance only by joint training with mix batches.

To alleviate this imbalance issue in OCIL, existing replay-based methods incorporate their additional techniques, generally falling into two categories. The first category is \textit{training distinction} \cite{SSIL/AhnKLBKM21,ACE/CacciaAATPB22,AFS/LiangCCJZ24,MIR/AljundiBTCCLP19,ASER/ShimMJSKJ21,GSS/AljundiLGB19,LODE/LiangL23}, which imposes different training policies or losses on buffer samples and new samples, respectively. Most works in this category focus on how to avoid the trained model being biased toward new classes by loss separation \cite{SSIL/AhnKLBKM21,ACE/CacciaAATPB22,LODE/LiangL23}. However, this conversely tends to reduce the learning performance, as the conflicting tasks of learning new classes and replaying buffer samples are \textit{competitively} and \textit{exclusively} performed within a single classifier, without complementing each other. The second category is \textit{feature enhancement} \cite{iCaRL/RebuffiKSL17,SCR/MaiLKS21,PCR/LinZFLY23,OCM/GuoLZ22,OnPro/0001YHZS23} that aims to obtain a more discriminative feature space where all embeddings, corresponding to either old classes or new classes, are well separated for better classification. This is typically done by using contrastive loss \cite{SCR/MaiLKS21}, as a replacement of cross-entropy loss, and by employing a post-hoc classifier, such as the nearest-class-mean (NCM) classifier \cite{NCM/MensinkVPC13}, which takes the best use of well-separated features. However, in the OCIL setting with a limited number of samples, it is challenging to maintain historical feature information, as there are fewer pairs available for effective contrastive learning.

To overcome these limitations, this paper proposes a novel replay-based OCIL method, called \textbf{B}alanced \textbf{I}nclusive \textbf{S}eparation for \textbf{O}nline i\textbf{N}cremental learning (BISON), designed to achieve a balanced yet satisfactory performance. Rather than competitively applying training policies or loss functions, our strategy is to incorporate separated components within the model itself, allowing for \textit{inclusive separation} yet clearer differentiation across the tasks of knowledge acquisition and preservation. Specifically, we employ dual classifiers with identical structures: the \textit{stream classifier}, which is dedicated to learning from new samples, and the \textit{buffer classifier}, which focuses on preserving existing knowledge using buffered samples. This structural separation not only enables each classifier to concentrate on its corresponding task, but also facilitates learning knowledge across old and new classes, thereby enhancing knowledge acquisition and consolidation.

For transferring knowledge across tasks, we introduce \textit{implicit} techniques that enable effective knowledge exchange between the dual classifiers, beyond simply blending old and new samples. Our first approach is redesigning \textit{proxy-anchor loss} (PAL) \cite{PAL/KimKCK20}, so that the weights of the stream classifier are treated as if they are learnable proxies during training. This method supports forward transfer by leveraging the previous knowledge of the feature extractor to guide the training of the stream classifier. For backward transfer, we employ a prototype-level proxy alignment feedback module that gradually transfers adaptive information from the stream classifier to the buffer classifier. With these techniques, both classifiers implicitly exchange and integrate their acquired knowledge.

As summarized in Figure~\ref{fig:intro_balance},
our empirical study shows that the proposed BISON method not only outperforms state-of-the-art replay-based OCIL methods in terms of overall performance (average accuracy), but also clearly achieves the best balance between plasticity (average intransigence) and stability (average forgetting).

\begin{figure*}[t]
\centering
\includegraphics[width=0.82\textwidth]{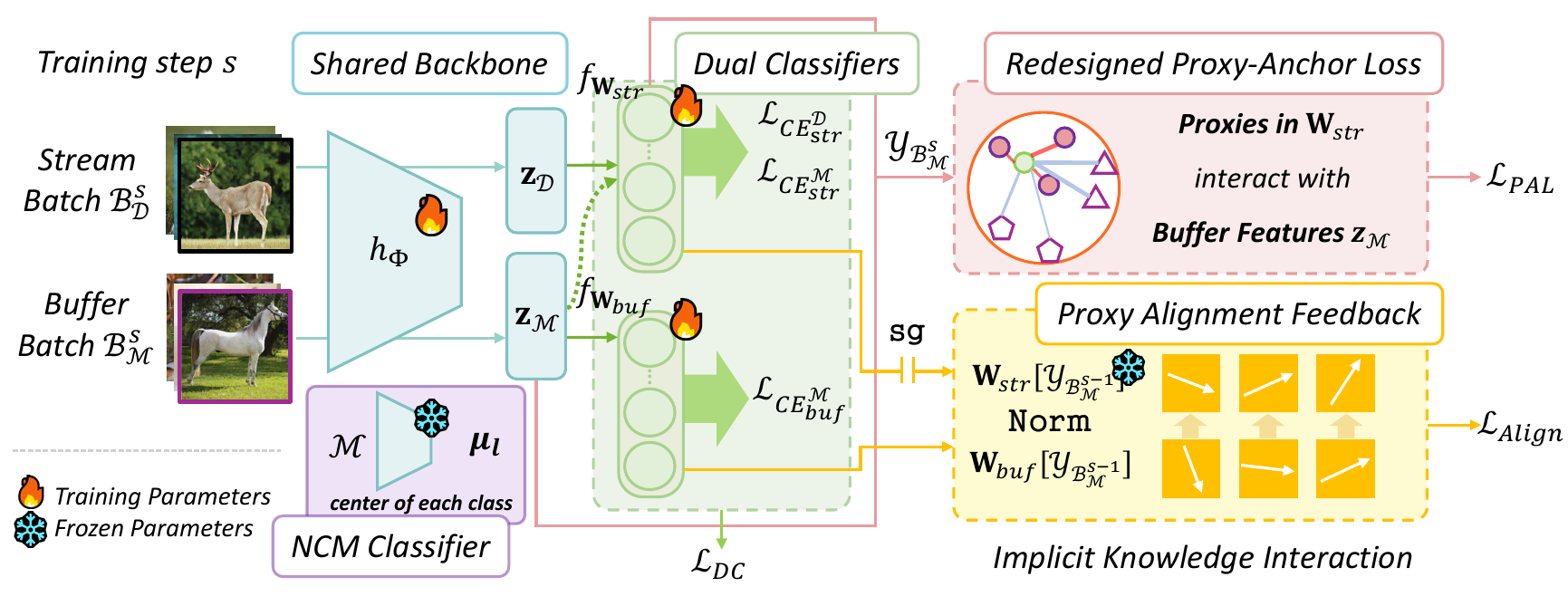}
\caption{Overview of our BISON method. For each batch of stream and buffer samples, the shared backbone outputs corresponding features $\mathbf{z}_{\mathcal{D}}, \mathbf{z}_{\mathcal{M}}$, which are then respectively fed into either stream or buffer classifier $\mathbf{W}_{str/buf}$ for separated training. Our redesigned proxy-anchor loss $\mathcal{L}_{PAL}$ and proxy alignment feedback $\mathcal{L}_{Align}$ are used for implicit knowledge exchange, helping both forward and backward knowledge transfer and better discrimination ability. Arrows of different colors represent flow directions participating in different loss modules. Based on the labels of the buffer batch in the previous step, we only align the proxies of the corresponding classes in the buffer classifier to the stream classifier using stop-gradient (denoted by $\mathtt{sg}$). We use the whole buffer $\mathcal{M}$ to acquire the center of each class and the NCM classifier is applied for inference.}
\label{fig:method_framework}
\end{figure*}

\section{Related Works} \label{sec:related}
Both online and offline continual learning (CL) typically focus on three major learning scenarios: class-incremental learning \cite{WA/ZhaoXGZX20,DERPP/BuzzegaBPAC20,DER/YanX021}, task-incremental learning \cite{DERPP/BuzzegaBPAC20}, and domain-incremental learning \cite{three/VenTT22}. This paper specifically deals with online class-incremental learning (OCIL), which is practically appealing for real-world applications but more challenging due to the streaming nature of data.

\smalltitle{Experience Replay}
Similar to the widely-used rehearsal method \cite{GEM/Lopez-PazR17,iCaRL/RebuffiKSL17,MER/RiemerCALRTT19,DERPP/BuzzegaBPAC20} in offline CL, \textit{experience replay} (ER) \cite{ER/chaudhry2019tiny} is considered the most effective approach in OCIL. The baseline ER method utilizes a bounded memory buffer to store and replay historical samples while jointly learning from new samples. As ER inevitably encounters imbalance issues between the streaming and buffered samples due to the bounded buffer, existing OCIL methods have introduced additional techniques, such as \textit{training distinction} and \textit{feature enhancement}.

\smalltitle{Training Distinction}
A number of existing works \cite{MIR/AljundiBTCCLP19,SSIL/AhnKLBKM21,DVC/Gu0WD22,ACE/CacciaAATPB22,LODE/LiangL23} can be classified as training distinction, which intends to apply different training policies or loss functions to buffer samples and stream samples to re-balance their contributions. For instance, \cite{SSIL/AhnKLBKM21} introduces loss separation, where losses for old and new classes are computed and backpropagated exclusively using a separated softmax function. Similarly, \cite{ACE/CacciaAATPB22} proposes an asymmetric cross-entropy loss to avoid the drastic drift in old features caused by new incoming batches. While these approaches can help address imbalance issues mostly by prioritizing previous samples, their exclusive separation scheme within a single classifier can hinder the model's ability to learn new knowledge as well as knowledge between old and new classes. Recently, \cite{LODE/LiangL23} attempts to smooth this exclusive separation by carefully decoupling the cross-entropy loss for stream samples into two terms, one for only new classes and the other for old and new classes, while all the loss terms are still applied to a single classifier. 

\smalltitle{Feature Enhancement}
Another group \cite{SCR/MaiLKS21,OCM/GuoLZ22,PCR/LinZFLY23,OnPro/0001YHZS23} focuses on feature enhancement, aiming to construct a more discriminative feature space by correcting or discarding biased features learned through cross-entropy loss. \cite{SCR/MaiLKS21} leverages supervised contrastive loss for representation learning and employs the NCM classifier at inference time. \cite{PCR/LinZFLY23} explores the coupling of proxy-based and contrastive-based loss by replacing anchor samples with proxies in contrastive loss. \cite{OCM/GuoLZ22} introduces mutual information maximization with InfoNCE loss to overcome catastrophic forgetting in online CL. Similarly, \cite{OnPro/0001YHZS23} utilizes InfoNCE but introduces online prototype learning to obtain representative features. However, compared to cross-entropy loss, contrastive learning typically requires extensive sample comparisons, which is challenging in OCIL with the bounded buffer.

\section{Problem Statement} \label{sec:problem}
\smalltitle{Problem Setting}
In OCIL, we are given a sequence of tasks $\mathcal{D}=\{\mathcal{D}_t\}_{t=1}^T$ from a single-pass data stream. Each task $t$ is associated with its dataset $\mathcal{D}_t = \{ \mathcal{X}_t \times \mathcal{Y}_t \}$, where $\mathcal{X}_t$ and $\mathcal{Y}_t$ represent samples and corresponding labels, respectively.
Without overlapping classes across tasks, each task $t$ corresponds to a set $C_t$ of classes such that $|C_t|$ is the same for all tasks. For the replay method, we also maintain a bounded memory buffer $\mathcal{M}$ that stores some of the previously trained samples, which is therefore updated over the learning steps. At each learning step $s$, the learner can access only a mini-batch $\mathcal{B}^s_{\mathcal{D}} \cup \mathcal{B}^s_{\mathcal{M}}$, where $\mathcal{B}^s_{\mathcal{D}}$ is a stream batch incoming from $\mathcal{D}$ and $\mathcal{B}^s_{\mathcal{M}}$ is a buffer batch retrieved from $\mathcal{M}$. Throughout the entire process, every instance in $\mathcal{D}$ can belong to only one stream batch $\mathcal{B}_{\mathcal{D}}$, and each $\mathcal{B}_{\mathcal{D}}$ can be trained only once.

\smalltitle{Baseline OCIL}
The general neural network architecture $\Theta = \{\Phi, \mathbf{W}\}$ for OCIL consists of a feature extractor $h$ and a classifier $f$, parameterized by $\Phi$ and $\mathbf{W}$, respectively. For each input $\mathbf{x}$, we predict its label based on the corresponding logit vector $\mathbf{s} = f(\mathbf{z}; \mathbf{W})$, where $\mathbf{z}=h(\mathbf{x};\Phi)$ is its feature embedding. Given a data stream $\mathcal{D}$ containing $T$ tasks and a target neural network $\Theta$, the baseline replay-based OCIL aims to continuously train $\Theta$ with each $\mathcal{D}_t$ for $t \in [1, T]$ such that $\Theta$ can make precise predictions for all the classes in $T$ tasks, by optimizing the following objective function:
\begin{equation} 
    \underset{\Phi, \mathbf{W}}{\arg \min } \enspace \mathbb{E}_{(\mathbf{x}, y) \sim \mathcal{D}_{t} \cup \mathcal{M}}\left[\mathcal{L}_{CE}\left(y, f(h(\mathbf{x};\Phi);\mathbf{W})\right)\right]. \nonumber
\end{equation}
$\mathcal{L}_{CE}\left(y, \mathbf{s}\right) = -\log \frac{\exp(\mathbf{s}^{(y)})}{\sum_{j\in C_{1:t}} \exp(\mathbf{s}^{(j)})}$ is cross-entropy loss, where $C_{1:t} = \cup_{i=1}^{t} C_t$ indicates all the classes learned so far and $\mathbf{s}^{(j)}$ is the individual logit value of class $j$.

\section{Methodology} \label{sec:method}
In this section, we present our proposed BISON method, which consists of two major components: (i) training separation with dual classifiers and (ii) implicit knowledge interaction. Intuitively, our techniques are motivated by both training distinction and feature enhancement, while introducing our novel approaches to address their limitations. The overall framework is illustrated in Figure \ref{fig:method_framework}.

\begin{figure}[t]
\centering
\subfigure[\label{fig:sep:a}]{\hspace{-3mm}\includegraphics[height=29mm]{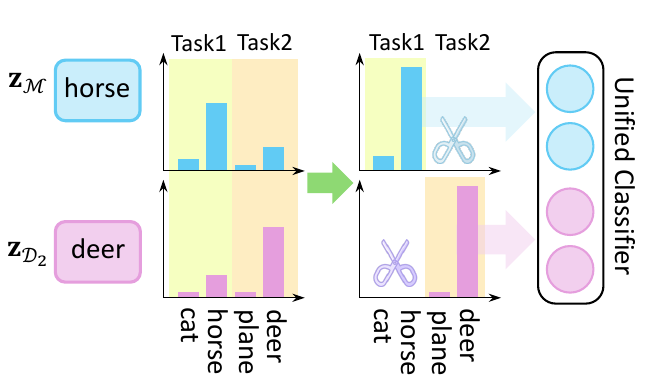}}
\subfigure[\label{fig:sep:b}]{\includegraphics[height=29mm]{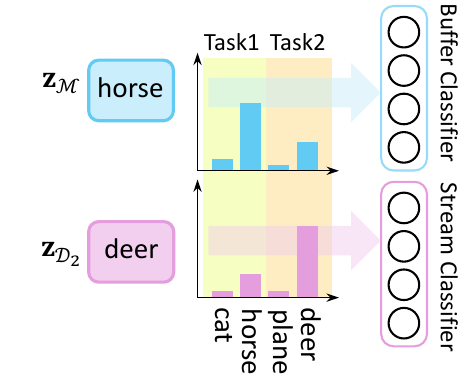}}
\vspace{-2mm}
\caption{\textbf{Comparison between two training separation schemes.} (a) Exclusive separation \cite{SSIL/AhnKLBKM21} within a single classifier. (b) Inclusive separation with dual classifiers.}
\label{fig:sep}
\vspace{-4mm}
\end{figure}

\subsection{Training Separation with Dual Classifiers}

The major obstacle of previous training distinction approaches is that two exclusive training schemes must compete in a single classifier. As illustrated in Figure \ref{fig:sep:a}, since both a buffer sample (e.g., a \textit{horse} image) and a new sample (e.g., a \textit{deer} image) share the same classifier, training separation must be exclusive in a way that each loss corresponds to either old classes or new classes, making it challenging to learn knowledge across tasks (e.g., relationship between \textit{horse} and \textit{deer}). Yet, without training distinction, this unified classifier can quickly get biased toward new classes. 

\smalltitle{Dual Classifiers for Training}
Based on the above motivation, our method proposes dual classifiers (DC) with the same structure, replacing the single classifier. These two independent classifiers, referred to as the stream classifier and the buffer classifier, are dedicated to the conflicting tasks of knowledge acquisition and preservation, respectively. Specifically, the stream classifier, parameterized by $\mathbf{W}_{str}$, primarily processes incoming batches (i.e., $\mathcal{B}_{\mathcal{D}}$) from the data stream, whereas the buffer classifier, parameterized by $\mathbf{W}_{buf}$, is fully responsible for learning from buffer batches (i.e., $\mathcal{B}_{\mathcal{M}}$). For the structure of each classifier, we adopt the idea of using cosine normalization \cite{Cosine/LuoZXWRY18,Cosine/HouPLWL19} that can eliminate the bias caused by difference in magnitudes, thereby alleviating imbalance. Formally, the logit for the $j$-th class is defined as:

\begin{equation} \label{eq:cosine_classifier_angle_bracket}
f_j(\mathbf{z}; \mathbf{W}) = \eta \cdot \text{cos}(\mathbf{W}_j, \mathbf{z})
\end{equation}
where $f_j(\mathbf{z}; \mathbf{W})$ is the resulting logit $s^{(j)}$ for the $j$-th class, $\mathbf{z}$ is the input feature vector, and $\mathbf{W}_j$ is the weight vector corresponding to the $j$-th class (i.e., the $j$-th row of the weight matrix $\mathbf{W} \in \mathbb{R}^{C \times D}$). The term $\eta$ is a trainable scaling factor, and the operator $\text{cos}(\mathbf{a},\mathbf{b})$ denotes the cosine similarity between two vectors $\mathbf{a}$ and $\mathbf{b}$, which is computed as $\frac{\mathbf{a}^T \mathbf{b}}{\|\mathbf{a}\| \|\mathbf{b}\|}$. As shown in Figure \ref{fig:method_framework}, both classifiers still share the feature extractor, from which we obtain the feature representations $\mathbf{z}_{\mathcal{D}}$ of incoming batches and those $\mathbf{z}_{\mathcal{M}}$ of buffer batches. Then, as indicated by the green solid arrows in Figure \ref{fig:method_framework}, $\mathbf{z}_{\mathcal{D}}$ and $\mathbf{z}_{\mathcal{M}}$ are fed into their respective classifiers to compute the logits and cross-entropy losses. 

This structural separation clearly allows the independent training of two tasks of knowledge acquisition and preservation without interference. Moreover, as illustrated in Figure \ref{fig:sep:b}, each task can still incorporate logit values from both old classes (e.g., \textit{cat} and \textit{horse}) and new classes (e.g., \textit{plane} and \textit{deer}), thereby facilitating the learning of relational knowledge (e.g., high relevance between \textit{deer} and \textit{horse}) across all classes. Such relationship information is particularly important for the feature extractor, which is shared by two classifiers, to learn better representation, aligning with the objective of feature enhancement.

\smalltitle{Separation Smoothing}
With dual classifiers, we define the loss function $\mathcal{L}_{DC}$ to train all of $\Phi$, $\mathbf{W}_{buf}$, and $\mathbf{W}_{str}$:
\begin{equation}
    \mathcal{L}_{DC} = \mathcal{L}_{CE_{str}^{\mathcal{D}}} + \alpha \mathcal{L}_{CE_{str}^{\mathcal{M}}} + (1-\alpha) \mathcal{L}_{CE_{buf}^{\mathcal{M}}},
\label{eq:dc}
\end{equation}
where the subscript $str$ or $buf$ refers to the stream or buffer classifier, and the superscript $\mathcal{D}$ or $\mathcal{M}$ indicates whether the loss is calculated based on the stream features $\mathbf{z}_{\mathcal{D}}$ or buffer features $\mathbf{z}_{\mathcal{M}}$. The adaptive separation smoother $\alpha$ is designed as a task-specific learnable parameter, initialized anew for each task and processed via a sigmoid. Note that the $\alpha \in [0,1]$ and when $\alpha$ is closer to $0$, it indicates stronger separation. This separation smoothing allows some buffer features to be fed into the stream classifier as well as the buffer classifier, which is indicated by the green dotted arrow in Figure \ref{fig:method_framework}. However, even in this smoother version, the buffer classifier is not designed to handle stream features for securing high stability.

Notably, our $\mathcal{L}_{DC}$ loss is entirely based on cross-entropy loss, from which the shared feature extractor can effectively be trained without extensive sample comparisons with the help of dual classifiers. Thus, we still take the benefit of cross-entropy learning as in existing training distinction methods, while enabling to obtain an improved feature extractor as in feature enhancement methods.

\smalltitle{NCM Classifier for Inference}
After training the feature extractor with dual classifiers using $\mathcal{L}_{DC}$, we employ the NCM classifier for making inference. This is because NCM is known to be more robust against potential biases in trained classifiers, as noted by \cite{SCR/MaiLKS21}, as long as the feature extractor can generate well-discriminated feature embeddings. To be shown by our experimental results in Table \ref{tab:NCM_comparison}, our BISON method takes the best out of NCM possibly due to its well-trained feature extractor.

\subsection{Implicit Knowledge Interaction}
Although our $\mathcal{L}_{DC}$ loss provides a way of knowledge exchange by mixing stream samples and buffer samples when $\alpha \in (0, 1)$, relying solely on this mixing strategy cannot resolve the imbalance problem, as in the baseline ER method. Therefore, we propose implicit techniques that enable effective knowledge exchange between the buffer classifier and the stream classifier, namely (i) redesigning proxy-anchor loss to the training of the stream classifier along with the feature extractor, and (ii) proxy alignment feedback to the buffer classifier.

\smalltitle{Buffer-to-Stream: Redesigned Proxy-Anchor Loss}
In our separated learning scheme with DC, the stream classifier may not effectively acquire the knowledge of buffer samples, as it is designed to guarantee such independent learning without any interference. However, forward transfer is essential in OCIL, where each stream batch contains only a small number of samples. To this end, we suggest using our redesigned version of proxy-anchor loss (PAL) \cite{PAL/KimKCK20} for the stream classifier to implicitly utilize knowledge from buffer features. The main idea is to let each class vector of the stream classifier, denoted as $\mathbf{w} \in \mathbf{W}_{str}$, be a prototype (i.e., proxy) of the class during training with features of each buffer batch $\mathcal{B}_{\mathcal{M}}$ using proxy-anchor loss, replacing randomly initialized proxies in the vanilla proxy-anchor loss. We use $\mathcal{Z}_{\mathcal{M}}$ to denote the set of buffer features corresponding to $\mathcal{B}_{\mathcal{M}}$. For each $\mathcal{Z}_{\mathcal{M}}$, we define our redesigned proxy-anchor loss $\mathcal{L}_{PAL}$ as follows:
\begin{equation} \label{eq:proxy_anchor_loss}
\begin{aligned}
\mathcal{L}_{PAL} = & \frac{1}{|\mathbf{W}_{str}^+|} \sum_{\mathbf{w} \in \mathbf{W}_{str}^+} \log ( 1 + \sum_{\mathbf{z} \in \mathcal{Z}_{M_w}^+} e^{-\gamma \left( \text{cos} ( \mathbf{z}, \mathbf{w} ) - \delta \right)} ) \\
& + \frac{1}{|\mathbf{W}_{str}|} \sum_{\mathbf{w} \in \mathbf{W}_{str}} \log ( 1 + \sum_{\mathbf{z} \in \mathcal{Z}_{M_w}^-} e^{\gamma \left( \text{cos} ( \mathbf{z}, \mathbf{w} ) + \delta \right)} )
\end{aligned}
\end{equation}

where $\delta>0$ is a margin, $\gamma>0$ is a constant scaling factor, determining how strongly pulling or pushing embedding vectors, and $\text{cos} (\mathbf{z}, \mathbf{w} )$ represents the cosine similarity between $\mathbf{z}$ and $\mathbf{w}$. $\mathbf{W}^{+}_{str}$ denotes positive proxies corresponding to the classes in the buffer batch. For each proxy $\mathbf{w}$, the set of buffer features $\mathcal{Z}_\mathcal{M}$ is also divided into its positive and negative subsets,  $\mathcal{Z}_{\mathcal{M}_\mathbf{w}}^{+}$ and $\mathcal{Z}_{\mathcal{M}_\mathbf{w}}^{-}$, respectively.

By minimizing $\mathcal{L}_{PAL}$, the stream classifier can fully utilize the knowledge of buffer samples, without explicitly training them on it using cross-entropy loss. Furthermore, previously trained but potentially removed samples from the buffer can still interact with remaining ones through proxy anchors in the stream classifier. This not only improves the knowledge consolidation in the feature space, but also makes the stream classifier more discriminative.

\begin{table*}[t]
\caption{Average accuracy at the end of training on three datasets. The best scores are highlighted in \textbf{boldface}, while the runner-up scores are \underline{underlined}. `T' and `F' indicate two categories: \textit{training distinction} and \textit{feature enhancement}, respectively.}
\label{tab:baselines}
\centering
\renewcommand{\arraystretch}{0.95}
\resizebox{0.95\textwidth}{!}{%
\begin{tabular}{lccclccclccc}
\hline
\multirow{2}{*}{Method} & \multicolumn{3}{c}{\textbf{Split CIFAR-100}} &  & \multicolumn{3}{c}{\textbf{Split CIFAR-10}} &  & \multicolumn{3}{c}{\textbf{Split Mini-ImageNet}} \\ \cline{2-4} \cline{6-8} \cline{10-12} 
 & $\mathcal{M}$ = 1k & $\mathcal{M}$ = 2k & $\mathcal{M}$ = 5k &  & $\mathcal{M}$ = 0.2k & $\mathcal{M}$ = 0.5k & $\mathcal{M}$ = 1k &  & $\mathcal{M}$ = 1k & $\mathcal{M}$ = 2k & $\mathcal{M}$ = 5k \\ \cline{1-4} \cline{6-8} \cline{10-12} 
FINE-TUNE & \multicolumn{3}{c}{5.2 $_{\pm0.6}$} &  & \multicolumn{3}{c}{17.4 $_{\pm0.8}$} &  & \multicolumn{3}{c}{4.6 $_{\pm0.7}$} \\ \hline
ER & 16.5 $_{\pm0.6}$ & 19.7 $_{\pm1.0}$ & 20.4 $_{\pm1.9}$ &  & 38.1 $_{\pm4.5}$ & 42.8 $_{\pm5.4}$ & 46.9 $_{\pm5.2}$ &  & 14.2 $_{\pm1.3}$ & 16.1 $_{\pm1.2}$ & 14.3 $_{\pm2.4}$ \\
GSS (T) & 16.6 $_{\pm0.9}$ & 18.7 $_{\pm1.3}$ & 18.2 $_{\pm0.9}$ &  & 25.2 $_{\pm2.0}$ & 30.2 $_{\pm2.4}$ & 38.7 $_{\pm4.0}$ &  & 13.0 $_{\pm0.9}$ & 14.7 $_{\pm1.9}$ & 14.6 $_{\pm2.1}$ \\
MIR (T) & 17.9 $_{\pm0.9}$ & 20.3 $_{\pm1.4}$ & 20.2 $_{\pm1.8}$ &  & 37.2 $_{\pm3.3}$ & 43.7 $_{\pm4.4}$ & 46.3 $_{\pm3.6}$ &  & 15.2 $_{\pm0.7}$ & 16.1 $_{\pm1.5}$ & 16.8 $_{\pm2.1}$ \\
A-GEM (T) & 5.3 $_{\pm0.4}$ & 5.0 $_{\pm0.5}$ & 5.7 $_{\pm0.3}$ &  & 17.4 $_{\pm1.0}$ & 17.1 $_{\pm1.3}$ & 17.6 $_{\pm1.0}$ &  & 4.5 $_{\pm0.5}$ & 4.9 $_{\pm0.5}$ & 4.9 $_{\pm0.4}$ \\
Gdumb (T) & 10.8 $_{\pm0.6}$ & 16.7 $_{\pm0.5}$ & 29.2 $_{\pm0.8}$ &  & 28.7 $_{\pm1.8}$ & 37.4 $_{\pm1.8}$ & 45.0 $_{\pm1.3}$ &  & 9.2 $_{\pm0.5}$ & 15.7 $_{\pm0.4}$ & 27.2 $_{\pm1.6}$ \\
SCR (F) & 13.6 $_{\pm0.9}$ & 14.9 $_{\pm0.8}$ & 15.8 $_{\pm0.6}$ &  & 46.1 $_{\pm2.1}$ & 54.8 $_{\pm1.5}$ & 57.8 $_{\pm1.6}$ &  & 13.0 $_{\pm0.6}$ & 14.6 $_{\pm0.4}$ & 15.9 $_{\pm0.6}$ \\
ASER (T) & 19.2 $_{\pm0.7}$ & 21.9 $_{\pm0.9}$ & 25.5 $_{\pm1.4}$ &  & 30.4 $_{\pm2.4}$ & 36.0 $_{\pm3.4}$ & 44.5 $_{\pm2.8}$ &  & 14.6 $_{\pm1.2}$ & 16.5 $_{\pm0.8}$ & 20.1 $_{\pm1.1}$ \\
SS-IL (T) & 21.1 $_{\pm0.8}$ & 22.5 $_{\pm0.7}$ & 22.3 $_{\pm0.6}$ &  & 41.3 $_{\pm1.1}$ & 43.8 $_{\pm2.0}$ & 47.7 $_{\pm2.0}$ &  & 19.0 $_{\pm1.1}$ & 20.5 $_{\pm1.0}$ & 20.3 $_{\pm0.8}$ \\
ER-DVC (T) & 19.3 $_{\pm1.2}$ & 22.2 $_{\pm1.5}$ & 23.9 $_{\pm1.4}$ &  & 45.6 $_{\pm2.8}$ & 45.4 $_{\pm3.5}$ & 52.1 $_{\pm2.8}$ &  & 17.0 $_{\pm1.0}$ & 17.6 $_{\pm1.6}$ & 18.8 $_{\pm1.7}$ \\
ER-ACE (T) & 23.0 $_{\pm0.4}$ & 25.6 $_{\pm0.8}$ & 27.7 $_{\pm0.9}$ &  & 48.0 $_{\pm2.2}$ & 54.0 $_{\pm1.0}$ & 58.6 $_{\pm1.7}$ &  & 20.5 $_{\pm1.7}$ & 23.6 $_{\pm1.4}$ & 25.2 $_{\pm1.9}$ \\
OCM (T, F) & 15.5 $_{\pm0.8}$ & 17.6 $_{\pm0.7}$ & 18.2 $_{\pm0.6}$ &  & 40.7 $_{\pm2.3}$ & 46.9 $_{\pm3.5}$ & 51.6 $_{\pm3.2}$ &  & 13.4 $_{\pm0.6}$ & 15.1 $_{\pm1.0}$ & 16.6 $_{\pm0.7}$ \\
PCR (F) & \underline{24.6} $_{\pm0.7}$ & \underline{27.3} $_{\pm0.9}$ & \underline{29.6} $_{\pm0.9}$ &  & 50.6 $_{\pm1.6}$ & 54.3 $_{\pm0.9}$ & 58.2 $_{\pm2.6}$ &  & \underline{23.9} $_{\pm0.6}$ & \underline{26.7} $_{\pm0.7}$ & 27.3 $_{\pm0.8}$ \\ 
LODE (T) & 24.4 $_{\pm1.1}$ & 26.5 $_{\pm1.1}$ & 29.0 $_{\pm1.1}$ &  & \underline{51.1} $_{\pm2.1}$ & \underline{56.9} $_{\pm2.9}$ & \underline{59.2} $_{\pm1.7}$ &  & 22.1 $_{\pm0.7}$ & 25.3 $_{\pm1.0}$ & \underline{27.8} $_{\pm0.9}$ \\
\hline \rowcolor{black!10}
\textbf{BISON (ours)} & \textbf{26.3} $_{\pm1.0}$ & \textbf{30.0} $_{\pm0.7}$ & \textbf{32.8} $_{\pm0.9}$ &  & \textbf{52.5} $_{\pm0.7}$ & \textbf{58.0} $_{\pm1.0}$ & \textbf{61.2} $_{\pm0.6}$ &  & \textbf{24.2} $_{\pm0.4}$ & \textbf{26.9} $_{\pm0.6}$ & \textbf{28.7} $_{\pm0.3}$  \\ \hline
\end{tabular}%
}
\end{table*}

\smalltitle{Stream-to-Buffer: Proxy Alignment Feedback}
For knowledge transfer from the stream classifier to the buffer classifier, we focus on the fact that the weight vector corresponding to each class in both classifiers technically serves as the class prototype. Based on this insight, we suggest proxy alignment feedback (PAF), that transfers the prototype-level information enhanced by our redesigned proxy-anchor loss. Specifically, given the set of class labels $\mathcal{Y}_{B^{s-1}_\mathcal{M}}$ observed in the previous buffer batch, we align the corresponding proxy weights between two classifiers as shown below:

\begin{equation}
    \mathcal{L}_{Align} = \frac{1}{|\mathcal{Y}_{B^{s-1}_\mathcal{M}}|} 
   \sum_{c \in \mathcal{Y}_{B^{s-1}_\mathcal{M}}} 
   \Big( 1 - \cos\!\big( \mathbf{W}^{s}_{buf}[c,:],\; \mathbf{W}^{s-1}_{str}[c,:] \big) \Big),
\label{eq:Align}
\end{equation}
where $\text{cos} (\cdot, \cdot )$ symbolizes cosine similarity as mentioned before and $\mathbf{W}^{s-1}_{str}[c,:]$ is frozen. The refined weights in $\mathbf{W}_{str}$ from the previous step are treated as fixed teacher proxies and are aligned with their buffer counterparts $\mathbf{W}_{buf}$. Note that the $\mathcal{L}_{Align}$ is calculated before the current step’s back-propagation, both classifiers are evaluated at the same temporal state, while the gradient is restricted to $\mathbf{W}_{buf}$.

\subsection{The Overall Process of BISON}

The overall process of BISON is illustrated in Figure \ref{fig:method_framework}. BISON comprises a shared feature extractor $h$, dual classifiers $f(\mathbf{z};{\mathbf{W}_{str/buf}})$ and implicit knowledge interaction modules. Combining all our components, we present the final loss of our BISON method as follows:
\begin{equation}
    \mathcal{L}_{BISON} = \mathcal{L}_{DC} + \beta  \mathcal{L}_{PAL} + \lambda\mathcal{L}_{Align},
\label{eq:totalloss}
\end{equation}
where $\beta$ and $\lambda$ are hyper-parameters. BISON trains a balanced online class-incremental learner through widely used cross-entropy loss together with redesigned proxy-anchor loss and proxy alignment feedback. More detailed steps are presented in the Appendix.

\section{Experiments} \label{sec:experiment}

\subsection{Experimental Setup}

\smalltitle{Datasets}
We use three real-world benchmark datasets in image classification. Following \cite{ASER/ShimMJSKJ21,DVC/Gu0WD22}, \textbf{Split CIFAR-10} is constructed by splitting CIFAR-10 \cite{CIFAR/krizhevsky2009learning} into 5 disjoint tasks, 2 classes per task. Both \textbf{Split CIFAR-100} and \textbf{Split Mini-ImageNet} contain 10 disjoint tasks, 10 classes per task, by splitting CIFAR-100 \cite{CIFAR/krizhevsky2009learning} and Mini-ImageNet \cite{Mini/VinyalsBLKW16}, respectively.

\smalltitle{Baselines}
We compare our BISON method with the following replay-based methods in the OCIL setting: ER \cite{ER/chaudhry2019tiny}, GSS \cite{GSS/AljundiLGB19}, MIR \cite{MIR/AljundiBTCCLP19}, A-GEM \cite{AGEM/chaudhry2018efficient}, Gdumb \cite{Gdumb/PrabhuTD20}, ASER \cite{ASER/ShimMJSKJ21}, SS-IL \cite{SSIL/AhnKLBKM21}, ER-DVC \cite{DVC/Gu0WD22}, ER-ACE \cite{ACE/CacciaAATPB22}, OCM \cite{OCM/GuoLZ22}, PCR \cite{PCR/LinZFLY23} and LODE \cite{LODE/LiangL23}. We also include FINE-TUNE as a non-replay baseline for comparison.

\smalltitle{Evaluation Metrics}
For performance assessment, we use three metrics commonly used in OCIL \cite{ASER/ShimMJSKJ21,Interplay/ChaudhryDAT18,CPR/ChaHHCM21}: \textit{average accuracy} (AA), \textit{average forgetting} (AF), and \textit{average intransigence} (AI). Intuitively, the higher AA indicates better overall performance, yet the lower AF and AI imply better stability and better plasticity, respectively. To provide their formal definitions, we first let $a_{k,j}\in [0,1]$ to denote the task-wise classification accuracy for the $j$-th task after learning $k \geq j$ continual tasks. According to the single-head evaluation setup \cite{Interplay/ChaudhryDAT18}, each prediction is made across all classes without being aware of task identification. Then, at the $k$-th task, three metrics are defined as follows \cite{CPR/ChaHHCM21}:
$
AA_k=\frac{1}{k}\sum_{j=1}^{k}{a_{k,j}},~AF_k=\frac{1}{k-1}\sum_{j=1}^{k-1}{f_{j,k}},$ and $AI_k=\frac{1}{k}\sum_{j=1}^{k}{a_{j}^{*}}-a_{j,j},
$
where (i) $f_{j,k}=\operatorname{max}_{i\in\{1,\dots,k-1\}}(a_{i,j} - a_{k,j})$ for $\forall{j<k}$, and (ii) $a_{j}^{*}$ indicates an empirical upper-bound for the task-wise accuracy of the $j$-th task, which is obtained from a purely fine-tuned model with respect to $\mathcal{D}_j$ without using any other loss terms. In terms of their value ranges, we have $AA \in[0,1]$ , $AF \in[-1,1]$, and $AI \in[-1,1]$.

\smalltitle{Implementation Details}
Following the common architecture in OCIL \cite{SCR/MaiLKS21,DVC/Gu0WD22,PCR/LinZFLY23}, we utilize Reduced ResNet-18 \cite{ResNet/HeZRS16} as the feature extractor. For batch size, each stream batch contains 10 images drawn from the data stream, while 10 samples are randomly retrieved from the buffer to form a buffer batch. In our BISON method, we keep the default setting as \cite{PAL/KimKCK20} does where $\gamma=32$ and $\delta=0.1$ in Eq. (\ref{eq:totalloss}) for all datasets. 
For Split CIFAR-100 and Split Mini-ImageNet, $\beta$ and $\lambda$ in Eq. (\ref{eq:totalloss}) are set to $0.2$ and $10.0$, respectively, whereas they are set to $0.1$ and $3.0$ for Split CIFAR-10. We reproduce all the evaluations in a consistent environment, where NVIDIA Geforce 3090 GPU and PyTorch toolbox are utilized. Each measurement in all experimental results is the average along with its standard deviation over 10 independent runs, where each run shuffles classes when splitting datasets. Full implementation details and hyperparameter determination are presented in the Appendix.

\begin{figure*}[t]
	\centering
    \subfigure[\label{fig:exp_balance:a}{Split CIFAR-100 ($\mathcal{M}$ = 2k)} ]{\hspace{0mm}\includegraphics[width=0.3\textwidth]{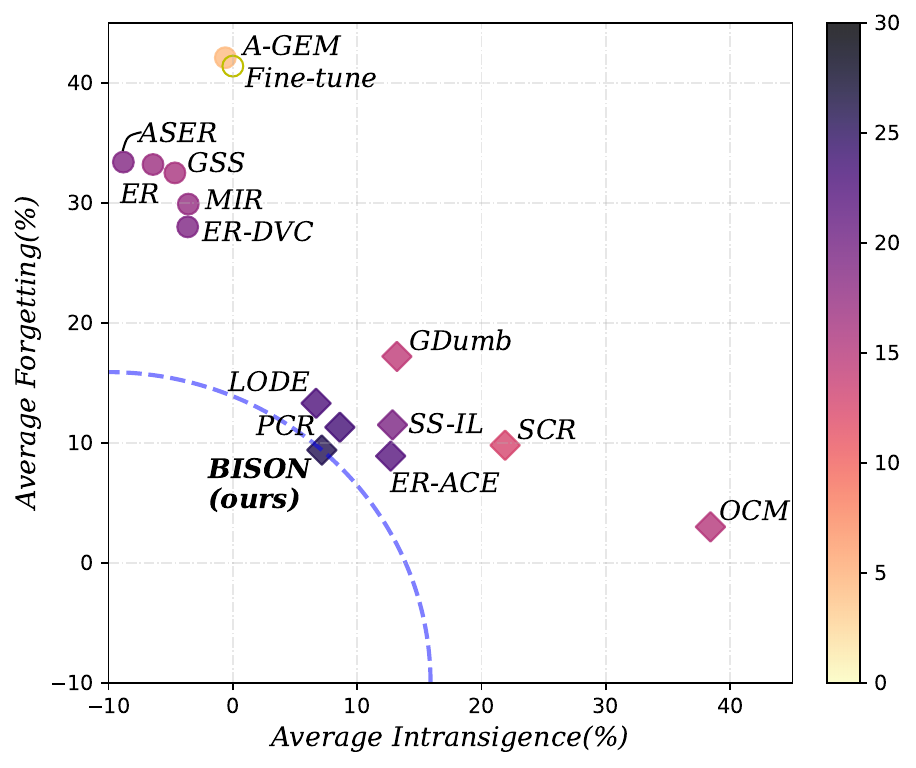}}
    \subfigure[\label{fig:exp_balance:b}{Split CIFAR-10 ($\mathcal{M}$ = 0.5k)} ]{\hspace{0mm}\includegraphics[width=0.3\textwidth]{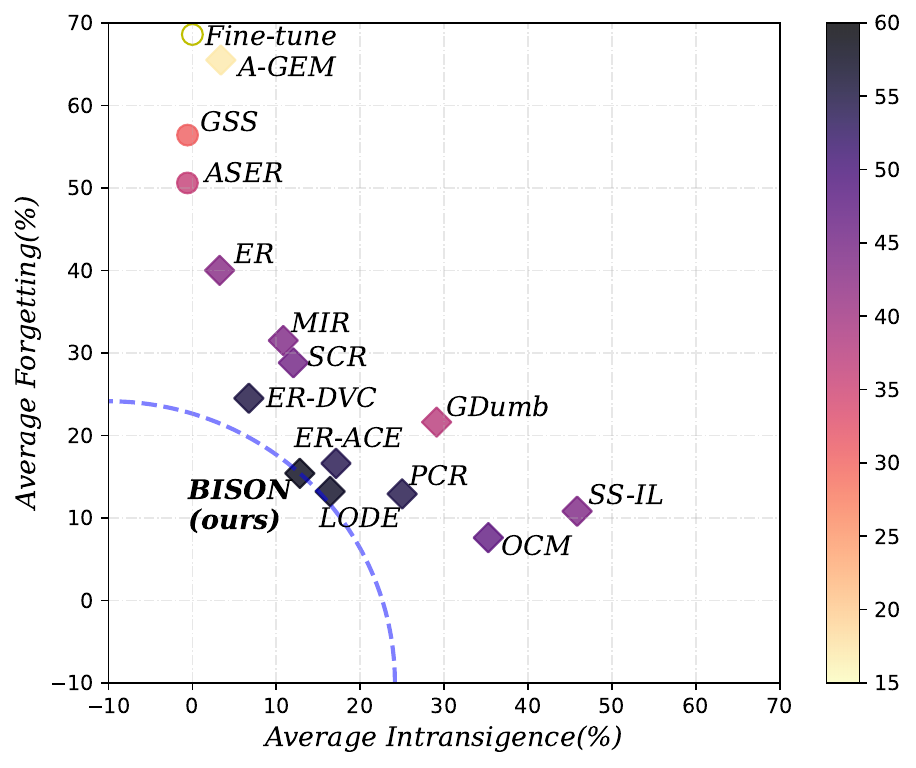}}
    \subfigure[\label{fig:exp_balance:c}{Split Mini-ImageNet ($\mathcal{M}$ = 2k)} ]{\hspace{0mm}\includegraphics[width=0.3\textwidth]{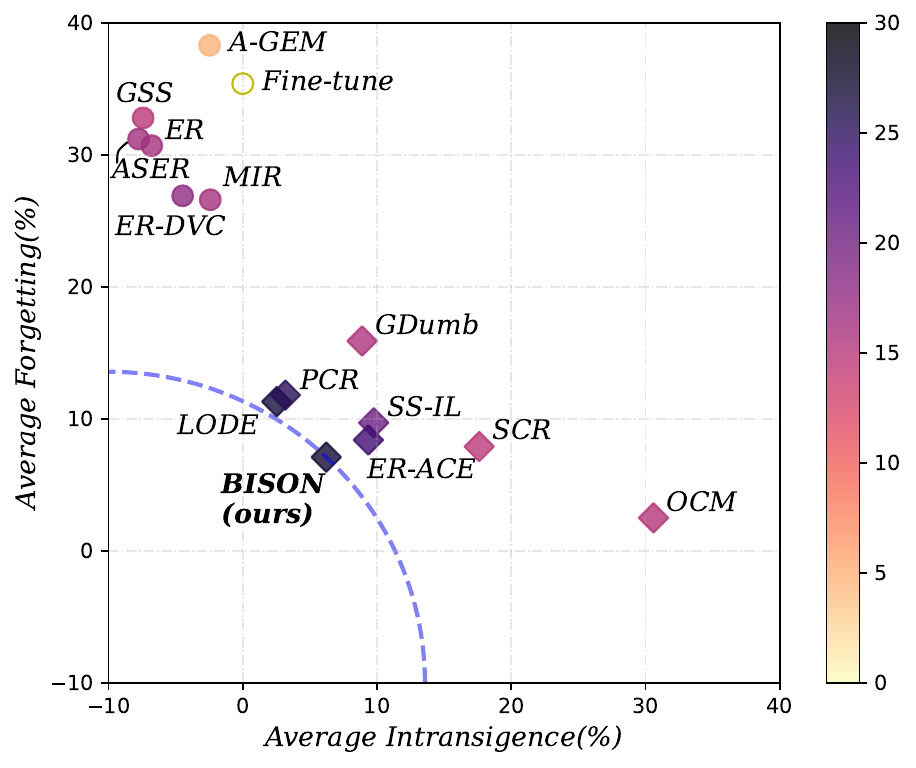}}
    \vspace{-2mm}
    \caption{\textbf{Balance between stability and plasticity.} Each graph plots average intransigence (AI) on the x-axis, which measures plasticity, and average forgetting (AF) on the y-axis, which measures stability. Both AI and AF are equally scaled in $[-1,1]$, with lower values indicating better plasticity and stability, respectively. Different colors represent levels of average accuracy.}
    \label{fig:exp_balance}
\end{figure*}

\subsection{Performance Comparison}

\smalltitle{Overall Performance}
Table \ref{tab:baselines} summarizes overall performance evaluation using three datasets, where we present the average accuracy of each model at the end of training over all tasks with different buffer sizes. Our BISON consistently achieves the best performance among all compared methods in all settings, mostly followed by PCR \cite{PCR/LinZFLY23}. Training distinction methods generally tend to show unsatisfactory performance with 
clear margins except for LODE \cite{LODE/LiangL23}, compared to BISON. Either SS-IL \cite{SSIL/AhnKLBKM21} or ER-ACE \cite{ACE/CacciaAATPB22} relies on exclusive loss separation, leading to less competitive performance, while LODE manages to further improve the performance by relaxed loss separation. In the category of feature enhancement, PCR takes the best position by leveraging proxies in metric learning, but SCR \cite{SCR/MaiLKS21} and OCM \cite{OCM/GuoLZ22} seem to suffer from too few samples in a mini-batch to train with their contrastive loss. We also examine the case of the larger buffer batch in the Appendix. Every method obviously shows better performance as buffer size increases, but our BISON method takes the best use of knowledge from buffer samples, further enlarging the performance gap with a larger buffer space. We also report the incremental performance over the steps in Figure \ref{fig:exp_curve} of Appendix, where our BISON method generally shows the highest performance throughout the entire training process.

\smalltitle{Balance of Stability and Plasticity}
To evaluate the balance between plasticity and stability, Figure \ref{fig:exp_balance} displays \textit{interplay graphs} \cite{Interplay/ChaudhryDAT18}, which plot each method based on two key metrics: average forgetting (AF) and average intransigence (AI). We also use different colors to show different levels of average accuracy (AA). Both AF and AI have the same range $[-1, 1]$, where lower values represent better stability and plasticity, respectively. Note that although AF and AI can technically be negative, a zero value already indicates no deviation from the ideal performance achievable through joint training with the entire dataset. In all the graphs, our BISON method is clearly positioned closest to the bottom-left corner, and is represented with the darkest color. This demonstrates that BISON not only outperforms exiting methods in terms of overall performance, but also consistently achieves the best balance between stability and plasticity. Methods like OCM \cite{OCM/GuoLZ22} tend to prioritize stability at the expense of plasticity. In contrast, methods with negative AI values (depicted as circular points), such as vanilla ER \cite{ER/chaudhry2019tiny}, excessively focus on plasticity, yet they suffer from severe forgetting with respect to AF, consequently failing to secure high AAs. Overall, maintaining balance is crucial for achieving high performance in OCIL, as more balanced methods tend to have darker-colored points, corresponding to higher AAs, with BISON leading the way, followed by LODE, PCR and ER-ACE.

\begin{table}[t]
\caption{Ablation study on Split Mini-ImageNet ($\mathcal{M}$=1k). ``w/o all'' represents vanilla ER with the NCM classifier.}
\centering
\renewcommand{\arraystretch}{1.0}
\resizebox{0.98\columnwidth}{!}{%
\begin{tabular}{lcccccc}
\hline
Method & DC & $\mathcal{L}_{PAL}$ & $\mathcal{L}_{Align}$ & AA $\uparrow$ & AF $\downarrow$ & AI $\downarrow$ \\ \hline
\textbf{BISON (ours)} & \cmark & \cmark & \cmark & $\bm{24.2\pm0.4}$ & $10.0\pm0.6$ & $\bm{5.9\pm8.7}$ \\ \hline
w/o DC \& PAF & \xmark & \cmark & \xmark & $18.1 \pm1.3$ & $\bm{7.4 \pm0.4}$ & $14.8 \pm8.4$ \\
w/o PAF \& PAL & \cmark & \xmark & \xmark & $22.6 \pm0.8$ & \multicolumn{1}{l}{$11.0 \pm0.7$} & $6.5 \pm7.5$ \\
w/o PAF & \cmark & \cmark & \xmark & $23.4 \pm0.6$ & $10.1 \pm0.8$ & $6.5 \pm9.1$ \\
w/o PAL & \cmark & \xmark & \cmark & $22.5 \pm0.6$ & $10.4 \pm1.0$ & $7.3 \pm8.2$ \\
w/o all & \xmark & \xmark & \xmark & $18.9 \pm1.1$ & $8.1 \pm0.9$ & $13.1 \pm8.4$ \\
\hline
\end{tabular}%
}
\label{tab:ablation}
\end{table}

\subsection{Ablation Studies}
\smalltitle{Impact of Each Component}
Table \ref{tab:ablation} presents the results of our ablation study on Split Mini-ImageNet with 1k buffer samples. To assess the impact of different components in the BISON method, we examine performance changes of every possible combinations\footnote{Note that PAF cannot solely be used without DC, and hence either `w/o DC' or `w/o DC \& PAL' is not a feasible case.} of individual components: Dual Classifiers (DC), redesigned Proxy-Anchor Loss (PAL), and Proxy Alignment Feedback (PAF), while commonly using the NCM classifier for inference. We first verify the impact of DC by removing DC and PAF together, as PAF is a backward transfer strategy that depends on DC and cannot function independently. The results, indicated by the reduced AA and increased AI in the `w/o DC \& PAF' setting, demonstrate the critical role of DC in enhancing plasticity as well as accuracy. This is partly because removing DC also includes discarding its underlying structure using cosine normalization. The comparison between BISON and `w/o PAF' shows that PAF further improves both plasticity and accuracy when combined with DC. Removing PAL also results in a decline in both stability and plasticity corresponding to increased AF and AI values in `w/o PAL', leading to a decrease in AA. The `w/o all' setting, which represents vanilla ER with NCM, exhibits the worst overall performance. This indicates that every component is essentially required to obtain the final performance and to maintain a good balance between AF and AI.

\begin{table}[t]
\caption{Average accuracy with or without using the NCM classifier on Split Mini-ImageNet with various buffer sizes.} 
\label{tab:NCM_comparison}
\centering
\renewcommand{\arraystretch}{1.0}
\resizebox{0.8\columnwidth}{!}{%
\begin{tabular}{lccc}
\hline
Method & $\mathcal{M}$ = 1k & $\mathcal{M}$ = 2k & $\mathcal{M}$ = 5k \\ 
\hline
ER & $14.2\pm1.3$ & $16.1\pm1.2$ & $14.3\pm2.4$ \\
ER + NCM & $18.9\pm1.1$ & $21.4\pm1.4$ & $20.9\pm2.2$ \\ \hline
SCR & $13.0\pm0.6$ & $14.6\pm0.4$ & $15.9\pm0.6$ \\ \hline
ER-ACE & $20.5\pm1.7$ & $23.6\pm1.4$ & $25.2\pm1.9$ \\
ER-ACE + NCM & $21.9\pm0.7$ & $25.1\pm0.6$ & $26.7\pm0.7$ \\ \hline
PCR & $23.9\pm0.6$ & $26.7\pm0.7$ & $27.3\pm0.8$ \\
PCR + NCM & $23.4\pm0.5$ & $26.0\pm0.5$ & $27.4\pm0.4$ \\ \hline
LODE & $22.1\pm0.7$ & $25.3\pm1.0$ & $27.8\pm0.9$ \\
LODE + NCM & $22.9\pm0.3$ & $26.3\pm0.5$ & $28.4\pm0.7$ \\ \hline
\textbf{BISON (ours)} & $\bm{24.2\pm0.4}$ & $\bm{26.9\pm0.6}$ & $\bm{28.7\pm0.3}$ \\ 
\hline
\end{tabular}%
}
\end{table}

\smalltitle{Impact of NCM}
From the result of the `w/o all' setting in Table \ref{tab:ablation}, the NCM classifier itself seems to be effective at enhancing stability even when used with vanilla ER. Motivated by this, in Table \ref{tab:NCM_comparison}, we further examine whether NCM can similarly enhance performance when combined with competitive state-of-the-art methods on Split Mini-ImageNet with varying buffer sizes. Note that the NCM classifier is already embedded in the SCR method as well as in our BISON method. As presented in Table \ref{tab:NCM_comparison}, our BISON method still outperforms all the compared methods regardless of whether they use NCM or not. While NCM generally improves classification accuracy, its impact varies depending on the underlying training methods. Specifically, vanilla ER takes the greatest benefit of using NCM, probably because the NCM classifier can mitigate class bias toward new tasks, which is a significant issue in the linear classifier of vanilla ER. In contrast, the performance improvement is less pronounced or occasionally negative in more advanced methods like ER-ACE and PCR. Notably, compared to SCR, which also utilizes NCM in its methodology, BISON substantially improves the performance with the help of our proposed techniques.

\begin{figure}[t!] 
\centering 
\begin{minipage}[h]{0.48\linewidth} \includegraphics[width=\linewidth]{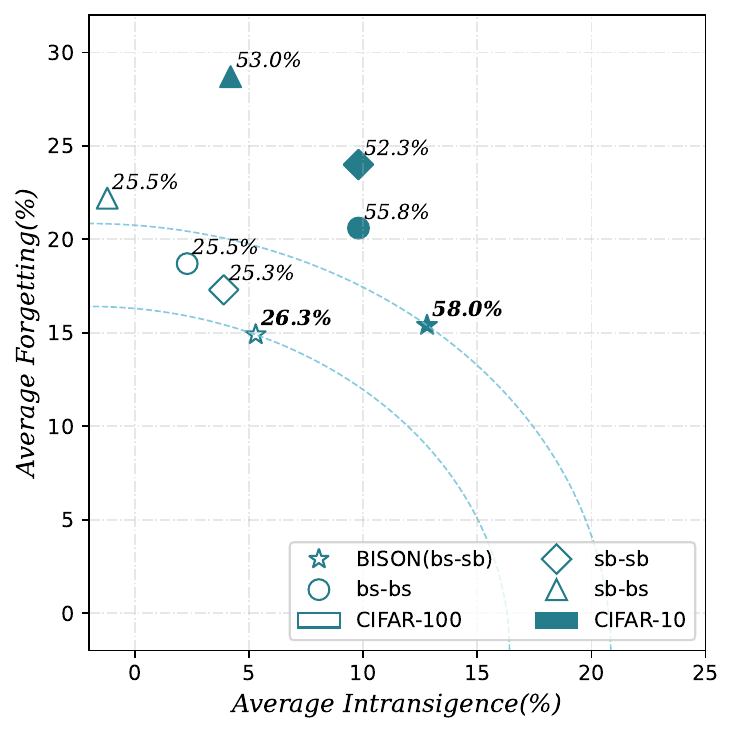} 
\caption{Performance under different directions of knowledge interaction.}  \label{fig:exp_dirc} \end{minipage} \hfill 
\hspace{1mm}
\begin{minipage}[h]{0.48\linewidth} \includegraphics[width=\linewidth]{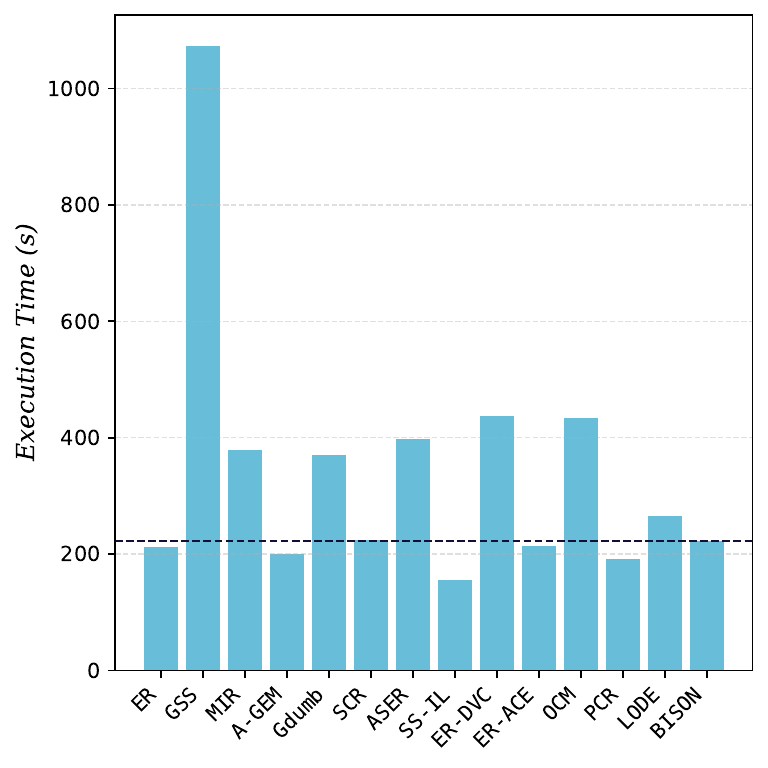}
\caption{Comparison of execution time on Split CIFAR-10 ($\mathcal{M}$ = 1k).}
\label{fig:exp_time} \end{minipage}
\vspace{-2mm}
\end{figure}

\smalltitle{Directions of Knowledge Interaction}
Figure~\ref{fig:exp_dirc} explores alternative directions of knowledge interaction between two classifiers, using Split CIFAR-100 ($\mathcal{M}$ = 1k) and Split CIFAR-10 ($\mathcal{M}$ = 0.5k). Recall that our proposed scheme, denoted as \textit{(bs, sb)}, employs PAL loss for forward transfer (i.e., buffer-to-stream, abbreviated as \textit{bs}), and  applies prototype-level PAF for backward transfer (i.e., stream-to-buffer, abbreviated as \textit{sb}). We additionally test three possible variants \textit{(bs, bs)}, \textit{(sb, sb)}, and \textit{(sb, bs)}, and compare them with our BISON method \textit{(bs, sb)}. As shown in Figure~\ref{fig:exp_dirc}, our proposed direction clearly yields the best AA by achieving an optimal balance between AF and AI. Interestingly, all other alternative directions tend to enhance plasticity but fail to preserve knowledge in the buffer, resulting in lower AAs than ours.

\smalltitle{Efficiency Analysis}
To study the practical feasibility of our proposed BISON method, we compare the execution times of all baselines on Split CIFAR-10 with a buffer size of 1k. The execution time encompasses both training and inference procedures. As shown in Figure~\ref{fig:exp_time}, the efficiency of BISON is comparable to ER, SCR, ER-ACE, PCR and LODE, while surpassing them in accuracy. In contrast, methods like GSS, MIR, ASER and ER-DVC, introduce their proposed strategies for buffer management or buffer sample retrieval, which incur significant computational overhead. Similarly, Gdumb involves relatively extensive training with buffer samples over several epochs, which is rare in typical online learning methods, and OCM utilizes rotation augmentation, which also adds to the processing time. Moreover, as presented in Table \ref{tab:comp_efficiency} of Appendix, BISON increases model size by only 0.1\% (CIFAR-10) and 1.4\% (CIFAR-100), with GPU usage increasing less than 1\%. Overall, our proposed method is efficient yet delivers the best prediction performance.

\begin{figure*}[t]
\centering
\includegraphics[height=58mm]{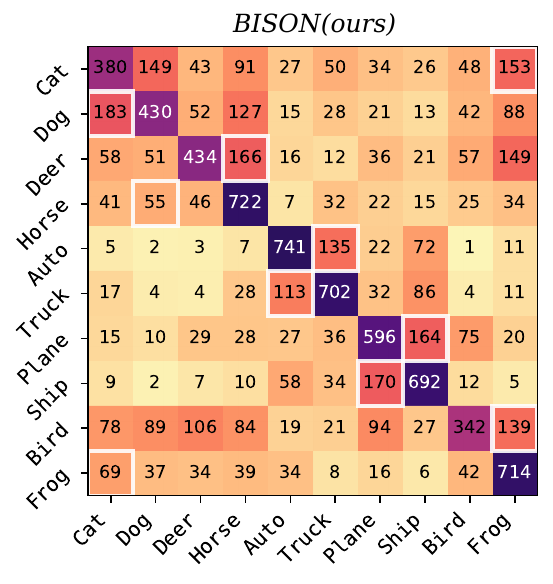} \hspace{3mm}
\includegraphics[height=58mm]{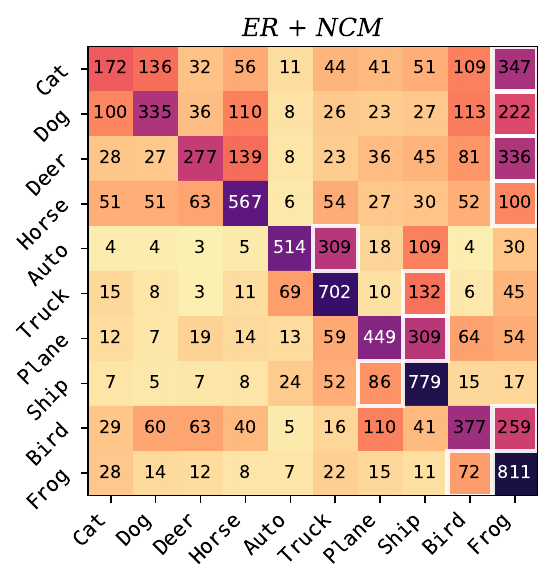} \hspace{3mm}
\includegraphics[height=58mm]{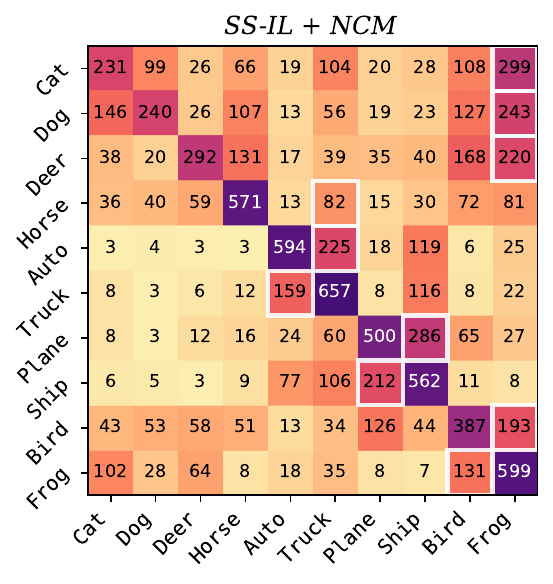}
\vspace{-2mm}
\caption{Confusion matrix on Split CIFAR-10 ($\mathcal{M}$ = 0.5k). The X-axis is prediction and the Y-axis is the label. Each off-diagonal white outline represents the top-1 misprediction for each class.}
\label{fig:confusion}
\end{figure*}

\subsection{Quantification of Cross-Task Confusion}

Although we have confirmed the effectiveness of each component in our ablation study shown in Table \ref{tab:ablation}, we further conduct an in-depth similar-pair analysis on the benefit of our design.

\smalltitle{Metrics} For any class $c$, we define the \textit{similar-neighbor} set $N(c)$ according to its semantic-similar pair. The global confusion matrix $M \in \mathbb{N}^{C \times C}$ is obtained after training all incremental tasks in a prefixed order and we normalize it along the row to get conditional prediction probabilities: $M_{\mathrm{row}}[c,d] = \mathrm{Pr}(\hat{y}=d |y=c)=\frac{M[c,d]}{\sum_j{M[c,j]}+\epsilon}$. Here, we formulate Similar-Confusion at top-1 ($\mathrm{SC@1}$) for per class as $\mathrm{SC@1}(c)=\sum_{d\in N(c)}M_{\mathrm{row}}{[c,d]}$, which reflects proportion of errors that specifically go to the similar neighbors. Moreover, we consider the precision for each class inside the similar pair ($P_{\mathrm{sim}}$), defined as $P_{\mathrm{sim}}(c)=\frac{M_{\mathrm{row}}[c,c]}{M_{\mathrm{row}}[c,c]+\sum_{d\in N(c)}M_{\mathrm{row}}[c,d]}$, measuring discriminability against its most confusable classes.

\smalltitle{Protocol} 
Focusing on the semantic proximity of class labels \cite{WordNet/Miller95} and human intuition \cite{HumanIntuition/PetersonBGR19}, we define five semantic-similar pairs for the CIFAR-10 dataset, which comprise \{(\textit{Cat}-\textit{Dog}), (\textit{Deer}-\textit{Horse}), (\textit{Automobile}-\textit{Truck}), (\textit{Airplane}-\textit{Ship}), (\textit{Bird}-\textit{Frog})\}. Then, to construct the incremental tasks, we distribute the classes from these similarity pairs across different tasks, introduced in the following order: \{\textit{Cat}, \textit{Deer}\}, \{\textit{Dog}, \textit{Automobile}\}, \{\textit{Horse}, \textit{Airplane}\}, \{\textit{Truck}, \textit{Bird}\}, and \{\textit{Ship}, \textit{Frog}\}. The results are obtained by making predictions against 10k test samples, 1k pre class after training the last task \{\textit{Ship}, \textit{Frog}\} on Split CIFAR-10 with 0.5k memory buffer samples and all the results are averaged over 3 runs.

\smalltitle{Reducing Bias Towards New Tasks}
Our first observation is that predictions for the newest classes, \textit{Ship} and \textit{Frog}, are reduced when applying our inclusive separation and knowledge interaction. To mitigate the effect of NCM itself on reducing task-recency bias, we apply NCM classifier on both ER and SS-IL during the inference phase. As shown in Figure \ref{fig:confusion}, for ER and SS-IL, most of the off-diagonal white outlines are located in the prediction columns of \textit{Frog} and \textit{Ship}, especially the former, where the predictions to \textit{Frog} are even more than the correct predictions to the label itself (\textit{Cat} and \textit{Deer} in ER and \textit{Cat} in SS-IL). Moreover, our proposed design enhances the precision of predictions for these new classes: the precision for \textit{Ship} improves from ER's $\frac{779}{1,534} = 50.8\%$ to $\frac{692}{1,122} = 61.7\%$, and for \textit{Frog} from ER's $\frac{811}{2,221} = 36.5\%$ to $\frac{714}{1,324} = 53.9\%$, whereas SS-IL achieves $\frac{562}{1,255} = 44.8\%$ and $\frac{599}{1,717} = 34.9\%$, respectively.

\begin{figure}[t]
    \centering
\includegraphics[width=0.98\linewidth]{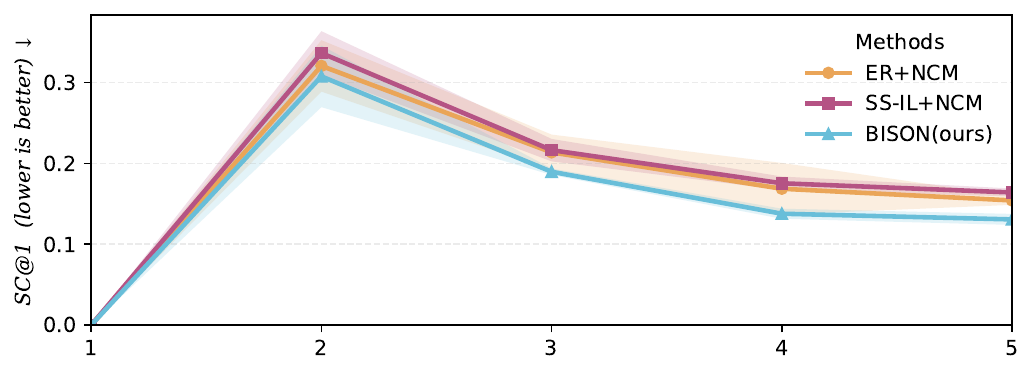}
\includegraphics[width=0.98\linewidth]{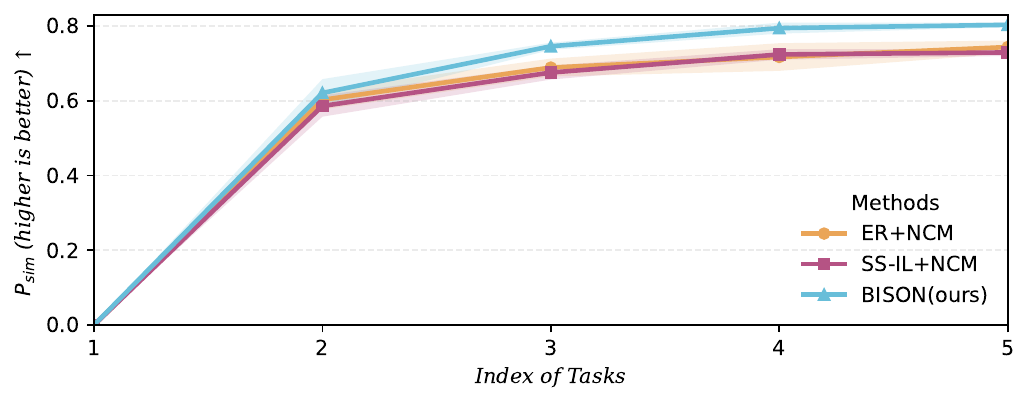}
\vspace{-2mm}
    \caption{Comparison of similar-confusion at top-1 ($\mathrm{SC}@1$) and precision for each class inside the similar-pair ($P_{sim}$) across tasks on Split CIFAR-10 ($\mathcal{M}$ = 0.5k).}
    \label{fig:sc1&psim}
\vspace{-2mm}
\end{figure}

\smalltitle{Capturing Semantic Relationship} Figure \ref{fig:confusion} also indicates that our inclusive separation with knowledge interaction bridges semantically similar classes and preserves cross-task semantic relations, as expected in Figure \ref{fig:sep}. We follow the main diagonal and focus on each block containing four cells formed by similar-pair. Our BISON method shows more white outlines inside the predefined similar-pair blocks (e.g., \textit{Cat}–\textit{Dog}, \textit{Deer}–\textit{Horse} and \textit{Automobile}-\textit{Truck}) than the baseline ER and SS-IL, indicating that residual errors remain within semantic neighbors rather than drifting to other classes. However, this increased correlation is what we intend to achieve for better feature embeddings, as it helps reduce incorrect predictions for unrelated classes, thereby enhancing overall accuracy. As presented in Figure \ref{fig:sc1&psim}, the mass of those confusions decreases ($\mathrm{SC}@1  \downarrow$) and the within-neighbor precision increases ($P_{\mathrm{sim}} \uparrow$ ) over tasks, consistently better than ER and SS-IL, i.e., the our BISON method incrementally preserves semantic structure and improves discrimination within the semantic similar-pair.

\section{Conclusion} \label{sec:conclusion}
We proposed a novel method for online class-incremental learning, named BISON, to overcome the limitations of existing replay-based methods, which often favor either plasticity or stability. BISON suggests employing dual classifiers to inclusively separate the learning of new samples from the replay of buffer samples, thereby mitigating bias toward new tasks yet effectively consolidating knowledge across different tasks. By introducing implicit knowledge exchange between the dual classifiers, BISON facilitates both forward and backward transfer over the incremental learning process. Empirical results show that BISON achieves the best performance by providing the most balanced learner for OCIL.

\begin{acks}
This work was supported in part by Institute of Information \& communications Technology Planning \& Evaluation (IITP) grants funded by the Korea government (MSIT) (No.2022-0-00448, Deep Total Recall: Continual Learning for Human-Like Recall of Artificial Neural Networks), and in part by INHA UNIVERSITY Research Grant.
\end{acks}

\printbibliography

@article{dilemma/mermillod2013stability,
  title={The stability-plasticity dilemma: Investigating the continuum from catastrophic forgetting to age-limited learning effects},
  author={Mermillod and
          Martial and 
          Bugaiska and
          Aur{\'e}lia 
          and Bonin, Patrick},
  journal={Frontiers in psychology},
  volume={4},
  pages={504},
  year={2013},
  publisher={Frontiers Media SA}
}

@article{AFS/LiangCCJZ24,
  author       = {Guoqiang Liang and
                  Zhaojie Chen and
                  Zhaoqiang Chen and
                  Shiyu Ji and
                  Yanning Zhang},
  title        = {New Insights on Relieving Task-Recency Bias for Online Class Incremental
                  Learning},
  journal      = {{IEEE} Trans. Circuits Syst. Video Technol.},
  volume       = {34},
  number       = {5},
  pages        = {3451--3464},
  year         = {2024},
  url          = {https://doi.org/10.1109/TCSVT.2023.3325651},
  doi          = {10.1109/TCSVT.2023.3325651},
}

@inproceedings{GEM/Lopez-PazR17,
  author       = {David Lopez{-}Paz and
                  Marc'Aurelio Ranzato},
  title        = {Gradient Episodic Memory for Continual Learning},
  booktitle    = {Advances in Neural Information Processing Systems 30: Annual Conference
                  on Neural Information Processing Systems 2017, December 4-9, 2017,
                  Long Beach, CA, {USA}},
  pages        = {6467--6476},
  year         = {2017},
}

@inproceedings{MER/RiemerCALRTT19,
  author       = {Matthew Riemer and
                  Ignacio Cases and
                  Robert Ajemian and
                  Miao Liu and
                  Irina Rish and
                  Yuhai Tu and
                  Gerald Tesauro},
  title        = {Learning to Learn without Forgetting by Maximizing Transfer and Minimizing
                  Interference},
  booktitle    = {7th International Conference on Learning Representations, {ICLR} 2019,
                  New Orleans, LA, USA, May 6-9, 2019},
  publisher    = {OpenReview.net},
  year         = {2019},
}

@inproceedings{Cosine/HouPLWL19,
  author       = {Saihui Hou and
                  Xinyu Pan and
                  Chen Change Loy and
                  Zilei Wang and
                  Dahua Lin},
  title        = {Learning a Unified Classifier Incrementally via Rebalancing},
  booktitle    = {{IEEE} Conference on Computer Vision and Pattern Recognition, {CVPR}
                  2019, Long Beach, CA, USA, June 16-20, 2019},
  pages        = {831--839},
  publisher    = {Computer Vision Foundation / {IEEE}},
  year         = {2019},
}

@inproceedings{DERPP/BuzzegaBPAC20,
  author       = {Pietro Buzzega and
                  Matteo Boschini and
                  Angelo Porrello and
                  Davide Abati and
                  Simone Calderara},
  title        = {Dark Experience for General Continual Learning: a Strong, Simple Baseline},
  booktitle    = {Advances in Neural Information Processing Systems 33: Annual Conference
                  on Neural Information Processing Systems 2020, NeurIPS 2020, December
                  6-12, 2020, virtual},
  year         = {2020},
}

@inproceedings{WA/ZhaoXGZX20,
  author       = {Bowen Zhao and
                  Xi Xiao and
                  Guojun Gan and
                  Bin Zhang and
                  Shu{-}Tao Xia},
  title        = {Maintaining Discrimination and Fairness in Class Incremental Learning},
  booktitle    = {2020 {IEEE/CVF} Conference on Computer Vision and Pattern Recognition,
                  {CVPR} 2020, Seattle, WA, USA, June 13-19, 2020},
  pages        = {13205--13214},
  publisher    = {Computer Vision Foundation / {IEEE}},
  year         = {2020},
}

@article{three/VenTT22,
  author       = {Gido M. van de Ven and
                  Tinne Tuytelaars and
                  Andreas S. Tolias},
  title        = {Three types of incremental learning},
  journal      = {Nat. Mac. Intell.},
  volume       = {4},
  number       = {12},
  pages        = {1185--1197},
  year         = {2022},
}

@inproceedings{Cosine/LuoZXWRY18,
  author       = {Chunjie Luo and
                  Jianfeng Zhan and
                  Xiaohe Xue and
                  Lei Wang and
                  Rui Ren and
                  Qiang Yang},
    title={Cosine normalization: Using cosine similarity instead of dot product in neural networks},
  author={Luo, Chunjie and Zhan, Jianfeng and Xue, Xiaohe and Wang, Lei and Ren, Rui and Yang, Qiang},
  booktitle={International conference on artificial neural networks},
  pages={382--391},
  year={2018},
  organization={Springer}
}

@inproceedings{ResNet/HeZRS16,
  author       = {Kaiming He and
                  Xiangyu Zhang and
                  Shaoqing Ren and
                  Jian Sun},
  title        = {Deep Residual Learning for Image Recognition},
  booktitle    = {2016 {IEEE} Conference on Computer Vision and Pattern Recognition,
                  {CVPR} 2016, Las Vegas, NV, USA, June 27-30, 2016},
  pages        = {770--778},
  publisher    = {{IEEE} Computer Society},
  year         = {2016},
}

@inproceedings{iCaRL/RebuffiKSL17,
  author       = {Sylvestre{-}Alvise Rebuffi and
                  Alexander Kolesnikov and
                  Georg Sperl and
                  Christoph H. Lampert},
  title        = {iCaRL: Incremental Classifier and Representation Learning},
  booktitle    = {2017 {IEEE} Conference on Computer Vision and Pattern Recognition,
                  {CVPR} 2017, Honolulu, HI, USA, July 21-26, 2017},
  pages        = {5533--5542},
  publisher    = {{IEEE} Computer Society},
  year         = {2017},
}

@inproceedings{DER/YanX021,
  author       = {Shipeng Yan and
                  Jiangwei Xie and
                  Xuming He},
  title        = {{DER:} Dynamically Expandable Representation for Class Incremental
                  Learning},
  booktitle    = {{IEEE} Conference on Computer Vision and Pattern Recognition, {CVPR}
                  2021, virtual, June 19-25, 2021},
  pages        = {3014--3023},
  publisher    = {Computer Vision Foundation / {IEEE}},
  year         = {2021},
}

@article{ER/chaudhry2019tiny,
  title={On tiny episodic memories in continual learning},
  author={Chaudhry, Arslan and Rohrbach, Marcus and Elhoseiny, Mohamed and Ajanthan, Thalaiyasingam and Dokania, Puneet K and Torr, Philip HS and Ranzato, Marc'Aurelio},
  journal={arXiv preprint arXiv:1902.10486},
  year={2019}
}

@inproceedings{GSS/AljundiLGB19,
  author       = {Rahaf Aljundi and
                  Min Lin and
                  Baptiste Goujaud and
                  Yoshua Bengio},
  title        = {Gradient based sample selection for online continual learning},
  booktitle    = {Advances in Neural Information Processing Systems 32: Annual Conference
                  on Neural Information Processing Systems 2019, NeurIPS 2019, December
                  8-14, 2019, Vancouver, BC, Canada},
  pages        = {11816--11825},
  year         = {2019},
}

@inproceedings{MIR/AljundiBTCCLP19,
  author       = {Rahaf Aljundi and
                  Eugene Belilovsky and
                  Tinne Tuytelaars and
                  Laurent Charlin and
                  Massimo Caccia and
                  Min Lin and
                  Lucas Page{-}Caccia},
  title        = {Online Continual Learning with Maximal Interfered Retrieval},
  booktitle    = {Advances in Neural Information Processing Systems 32: Annual Conference on Neural Information Processing Systems 2019, NeurIPS 2019, December
                  8-14, 2019, Vancouver, BC, Canada},
  pages        = {11849--11860},
  year         = {2019},
}

@inproceedings{AGEM/chaudhry2018efficient,
  author       = {Arslan Chaudhry and
                  Marc'Aurelio Ranzato and
                  Marcus Rohrbach and
                  Mohamed Elhoseiny},
  title        = {Efficient Lifelong Learning with {A-GEM}},
  booktitle    = {7th International Conference on Learning Representations, {ICLR} 2019,
                  New Orleans, LA, USA, May 6-9, 2019},
  publisher    = {OpenReview.net},
  year         = {2019},
}

@inproceedings{Gdumb/PrabhuTD20,
  author       = {Ameya Prabhu and
                  Philip H. S. Torr and
                  Puneet K. Dokania},
  title        = {GDumb: {A} Simple Approach that Questions Our Progress in Continual
                  Learning},
  booktitle    = {Computer Vision - {ECCV} 2020 - 16th European Conference, Glasgow,
                  UK, August 23-28, 2020, Proceedings, Part {II}},
  series       = {Lecture Notes in Computer Science},
  volume       = {12347},
  pages        = {524--540},
  publisher    = {Springer},
  year         = {2020},
}

@inproceedings{SCR/MaiLKS21,
  author       = {Zheda Mai and
                  Ruiwen Li and
                  Hyunwoo Kim and
                  Scott Sanner},
  title        = {Supervised Contrastive Replay: Revisiting the Nearest Class Mean Classifier
                  in Online Class-Incremental Continual Learning},
  booktitle    = {{IEEE} Conference on Computer Vision and Pattern Recognition Workshops,
                  {CVPR} Workshops 2021, virtual, June 19-25, 2021},
  pages        = {3589--3599},
  publisher    = {Computer Vision Foundation / {IEEE}},
  year         = {2021},
}

@inproceedings{ASER/ShimMJSKJ21,
  author       = {Dongsub Shim and
                  Zheda Mai and
                  Jihwan Jeong and
                  Scott Sanner and
                  Hyunwoo Kim and
                  Jongseong Jang},
  title        = {Online Class-Incremental Continual Learning with Adversarial Shapley
                  Value},
  booktitle    = {Thirty-Fifth {AAAI} Conference on Artificial Intelligence, {AAAI}
                  2021, Thirty-Third Conference on Innovative Applications of Artificial
                  Intelligence, {IAAI} 2021, The Eleventh Symposium on Educational Advances
                  in Artificial Intelligence, {EAAI} 2021, Virtual Event, February 2-9,
                  2021},
  pages        = {9630--9638},
  publisher    = {{AAAI} Press},
  year         = {2021},
}

@inproceedings{SSIL/AhnKLBKM21,
  author       = {Hongjoon Ahn and
                  Jihwan Kwak and
                  Subin Lim and
                  Hyeonsu Bang and
                  Hyojun Kim and
                  Taesup Moon},
  title        = {{SS-IL:} Separated Softmax for Incremental Learning},
  booktitle    = {2021 {IEEE/CVF} International Conference on Computer Vision, {ICCV}
                  2021, Montreal, QC, Canada, October 10-17, 2021},
  pages        = {824--833},
  publisher    = {{IEEE}},
  year         = {2021},
}

@inproceedings{DVC/Gu0WD22,
  author       = {Yanan Gu and
                  Xu Yang and
                  Kun Wei and
                  Cheng Deng},
  title        = {Not Just Selection, but Exploration: Online Class-Incremental Continual
                  Learning via Dual View Consistency},
  booktitle    = {{IEEE/CVF} Conference on Computer Vision and Pattern Recognition,
                  {CVPR} 2022, New Orleans, LA, USA, June 18-24, 2022},
  pages        = {7432--7441},
  publisher    = {{IEEE}},
  year         = {2022},
}

@inproceedings{ACE/CacciaAATPB22,
  author       = {Lucas Caccia and
                  Rahaf Aljundi and
                  Nader Asadi and
                  Tinne Tuytelaars and
                  Joelle Pineau and
                  Eugene Belilovsky},
  title        = {New Insights on Reducing Abrupt Representation Change in Online Continual
                  Learning},
  booktitle    = {The Tenth International Conference on Learning Representations, {ICLR}
                  2022, Virtual Event, April 25-29, 2022},
  publisher    = {OpenReview.net},
  year         = {2022},
}

@inproceedings{OCM/GuoLZ22,
  author       = {Yiduo Guo and
                  Bing Liu and
                  Dongyan Zhao},
  title        = {Online Continual Learning through Mutual Information Maximization},
  booktitle    = {International Conference on Machine Learning, {ICML} 2022, 17-23 July
                  2022, Baltimore, Maryland, {USA}},
  series       = {Proceedings of Machine Learning Research},
  volume       = {162},
  pages        = {8109--8126},
  publisher    = {{PMLR}},
  year         = {2022},
}

@inproceedings{PCR/LinZFLY23,
  author       = {Huiwei Lin and
                  Baoquan Zhang and
                  Shanshan Feng and
                  Xutao Li and
                  Yunming Ye},
  title        = {{PCR:} Proxy-Based Contrastive Replay for Online Class-Incremental
                  Continual Learning},
  booktitle    = {{IEEE/CVF} Conference on Computer Vision and Pattern Recognition,
                  {CVPR} 2023, Vancouver, BC, Canada, June 17-24, 2023},
  pages        = {24246--24255},
  publisher    = {{IEEE}},
  year         = {2023},
}

@article{NCM/MensinkVPC13,
  author       = {Thomas Mensink and
                  Jakob Verbeek and
                  Florent Perronnin and
                  Gabriela Csurka},
  title        = {Distance-Based Image Classification: Generalizing to New Classes at
                  Near-Zero Cost},
  journal      = {{IEEE} Trans. Pattern Anal. Mach. Intell.},
  volume       = {35},
  number       = {11},
  pages        = {2624--2637},
  year         = {2013},
}

@inproceedings{PAL/KimKCK20,
  author       = {Sungyeon Kim and
                  Dongwon Kim and
                  Minsu Cho and
                  Suha Kwak},
  title        = {Proxy Anchor Loss for Deep Metric Learning},
  booktitle    = {2020 {IEEE/CVF} Conference on Computer Vision and Pattern Recognition,
                  {CVPR} 2020, Seattle, WA, USA, June 13-19, 2020},
  pages        = {3235--3244},
  publisher    = {Computer Vision Foundation / {IEEE}},
  year         = {2020},
}

@inproceedings{Interplay/ChaudhryDAT18,
  author       = {Arslan Chaudhry and
                  Puneet Kumar Dokania and
                  Thalaiyasingam Ajanthan and
                  Philip H. S. Torr},
  title        = {Riemannian Walk for Incremental Learning: Understanding Forgetting
                  and Intransigence},
  booktitle    = {Computer Vision - {ECCV} 2018 - 15th European Conference, Munich,
                  Germany, September 8-14, 2018, Proceedings, Part {XI}},
  series       = {Lecture Notes in Computer Science},
  volume       = {11215},
  pages        = {556--572},
  publisher    = {Springer},
  year         = {2018},
}

@inproceedings{OnPro/0001YHZS23,
  author       = {Yujie Wei and
                  Jiaxin Ye and
                  Zhizhong Huang and
                  Junping Zhang and
                  Hongming Shan},
  title        = {Online Prototype Learning for Online Continual Learning},
  booktitle    = {{IEEE/CVF} International Conference on Computer Vision, {ICCV} 2023,
                  Paris, France, October 1-6, 2023},
  pages        = {18718--18728},
  publisher    = {{IEEE}},
  year         = {2023},
}

@article{CIFAR/krizhevsky2009learning,
  title={Learning multiple layers of features from tiny images},
  author={Krizhevsky, Alex and Hinton, Geoffrey and others},
  year={2009},
  publisher={Toronto, ON, Canada}
}

@inproceedings{Mini/VinyalsBLKW16,
  author       = {Oriol Vinyals and
                  Charles Blundell and
                  Tim Lillicrap and
                  Koray Kavukcuoglu and
                  Daan Wierstra},
  title        = {Matching Networks for One Shot Learning},
  booktitle    = {Advances in Neural Information Processing Systems 29: Annual Conference
                  on Neural Information Processing Systems 2016, December 5-10, 2016,
                  Barcelona, Spain},
  pages        = {3630--3638},
  year         = {2016},
}

@inproceedings{CPR/ChaHHCM21,
  author       = {Sungmin Cha and
                  Hsiang Hsu and
                  Taebaek Hwang and
                  Fl{\'{a}}vio P. Calmon and
                  Taesup Moon},
  title        = {{CPR:} Classifier-Projection Regularization for Continual Learning},
  booktitle    = {9th International Conference on Learning Representations, {ICLR} 2021,
                  Virtual Event, Austria, May 3-7, 2021},
  publisher    = {OpenReview.net},
  year         = {2021},
}

@inproceedings{LODE/LiangL23,
  author       = {Yan{-}Shuo Liang and
                  Wu{-}Jun Li},
  title        = {Loss Decoupling for Task-Agnostic Continual Learning},
  booktitle    = {Advances in Neural Information Processing Systems 36: Annual Conference
                  on Neural Information Processing Systems 2023, NeurIPS 2023, New Orleans,
                  LA, USA, December 10 - 16, 2023},
  year         = {2023},
}

@article{WordNet/Miller95,
  author       = {George A. Miller},
  title        = {WordNet: {A} Lexical Database for English},
  journal      = {Commun. {ACM}},
  volume       = {38},
  number       = {11},
  pages        = {39--41},
  year         = {1995},
  url          = {https://doi.org/10.1145/219717.219748},
  doi          = {10.1145/219717.219748},
}

@inproceedings{HumanIntuition/PetersonBGR19,
  author       = {Joshua C. Peterson and
                  Ruairidh M. Battleday and
                  Thomas L. Griffiths and
                  Olga Russakovsky},
  title        = {Human Uncertainty Makes Classification More Robust},
  booktitle    = {2019 {IEEE/CVF} International Conference on Computer Vision, {ICCV}
                  2019, Seoul, Korea (South), October 27 - November 2, 2019},
  pages        = {9616--9625},
  publisher    = {{IEEE}},
  year         = {2019},
  url          = {https://doi.org/10.1109/ICCV.2019.00971},
  doi          = {10.1109/ICCV.2019.00971},
}
\onecolumn
\setcounter{table}{0}
\setcounter{figure}{0}

\renewcommand{\thetable}{A\arabic{table}}
\renewcommand{\thefigure}{A\arabic{figure}}

\section*{Supplementary Material: Balanced Online Class-Incremental Learning via Dual Classifiers}
In this appendix, we first present a detailed pseudocode of our BISON method, which includes both training and inference processes. Then, we cover more comprehensive experimental results and full implementation details.

\subsection*{Algorithm of Balanced Inclusive
Separation for Online iNcremental learning (BISON)}
Algorithm \ref{alg:pdc:train} shows detailed steps on how BISON trains a neural network $\Theta$ with a given sequence of $T$ tasks, and how it makes a prediction for each test sample after incremental training. For each stream batch $\mathcal{B}^s_{\mathcal{D}_t}$ at the step $s$, we first randomly retrieve as many buffer samples as there are stream samples in $\mathcal{B}^s_{\mathcal{D}_t}$ (e.g., 10 samples) (Line 5), then apply data augmentation to both stream and buffer samples, thereby obtaining samples concatenated with their augmented counterparts (Lines 7--9). These original and augmented samples are then fed into the shared feature extractor $h(\cdot~; \Phi)$ to get their corresponding feature vectors (Lines 8--9), and each feature vector is accordingly assigned to either the stream classifier $f(\cdot~;\mathbf{W}_{str})$ or the buffer classifier $f(\cdot~;\mathbf{W}_{buf})$ for the computation of cross-entropy loss terms in Eq. (\ref{eq:dc}) (Lines 10--14). Additionally, the proxy-anchor loss is computed with respect to the buffer features $\mathbf{z}_{\mathcal{M}}$ and class proxies $\mathbf{W}_{str}$ in the stream classifier (Line 15). Based on previously stored label $y^{s-1}_{\mathcal{M}}$, the loss $\mathcal{L}_{Align}$ is calculated between corresponding rows in weights $\mathbf{W}_{str}$ and $\mathbf{W}_{buf}$ (Line 16). The final loss $\mathcal{L}_{BISON}$ is computed and used to update the network parameters (Line 17). To incorporate each stream batch into the buffer, we employ a standard buffer management scheme using reservoir sampling (Line 18). After each training step with a mini-batch, we store the label of buffer batch (Line 21).

After training is completed for each $t$-th task, we perform a testing procedure with test samples from all observed classes in $C_{1:t}$. During inference, we utilize the NCM classifier, where class-wise centroids are computed based on all the samples currently in the buffer (Lines 23--27).

\begin{algorithm}
	\renewcommand{\algorithmicrequire}{\textbf{Input:}}
	\renewcommand{\algorithmicensure}{\textbf{Output:}}
	\caption{Balanced Inclusive
Separation for Online iNcremental learning}
	\label{alg:pdc:train}
	\begin{algorithmic}[1]
		\REQUIRE Dataset $\mathcal{D}$; Learning Rate $\lambda$; Adaptive separation smoother $\alpha$; PAL Coefficient $\beta$; PAF Coeffcient $\lambda$.
        \renewcommand{\algorithmicensure}{\textbf{Initialize:}}
        \ENSURE Memory Buffer $\mathcal{M} \xleftarrow{} \{ \}$; Network Parameters $\Theta=\{\Phi, \mathbf{W}_{str}, \mathbf{W}_{buf}, \eta_{str}, \eta_{buf}, \alpha \}$; Data Augmentation $\textsc{Aug}(\cdot)$.
        \FOR{$t=1$ to $T$}     
		\STATE  /* Training Phase: */ 
        \STATE Initialize $\alpha\xleftarrow{}0$
            \FOR{mini-batch $\mathcal{B}^s_{\mathcal{D}_t} \sim \mathcal{D}_t$} 
            \STATE $ \mathcal{B}^s_{\mathcal{M}} \leftarrow \textsc{RandomRetrieval}(\mathcal{M})$ s.t. $|\mathcal{B}^s_{\mathcal{M}}| = |\mathcal{B}^s_{\mathcal{D}_t}|$
            \FOR{$(\mathbf{x}^s_{\mathcal{D}_t}, y^s_{\mathcal{D}_t}) \in \mathcal{B}^s_{\mathcal{D}_t},~(\mathbf{x}^s_{\mathcal{M}}, y^s_{\mathcal{M}}) \in \mathcal{B}^s_{\mathcal{M}}$}
            
            \STATE $\mathbf{x}_{\mathcal{M}}^{aug}, \mathbf{x}_{\mathcal{D}_t}^{aug} \xleftarrow{} \textsc{Aug}(\mathbf{x}^s_{\mathcal{M}}), \textsc{Aug}(\mathbf{x}^s_{\mathcal{D}_t})$
            \STATE $\mathbf{z}_{\mathcal{D}} \xleftarrow{} h(\textsc{Concat}([\mathbf{x}^s_{\mathcal{D}_t}, \mathbf{x}_{\mathcal{D}_t}^{aug}]); \Phi)$
            \STATE $\mathbf{z}_{\mathcal{M}} \xleftarrow{} h(\textsc{Concat}([\mathbf{x}^s_{\mathcal{M}}, \mathbf{x}^{aug}_{\mathcal{M}}]); \Phi)$
            \STATE $\mathcal{L}_{CE_{str}^{\mathcal{D}}} \xleftarrow{} \textsc{CrossEntropy}(\textsc{Concat}([y^s_{\mathcal{D}_t}, y^s_{\mathcal{D}_t}]),  f(\mathbf{z}_{\mathcal{D}}; \mathbf{W}_{str}))$
            \STATE $\mathcal{L}_{CE_{str}^{\mathcal{M}}} \xleftarrow{} \textsc{CrossEntropy}(\textsc{Concat}([y^s_{\mathcal{M}}, y^s_{\mathcal{M}}]),  f(\mathbf{z}_{\mathcal{M}}; \mathbf{W}_{str}))$
            \STATE $\mathcal{L}_{CE_{buf}^{\mathcal{M}}} \xleftarrow{} \textsc{CrossEntropy}(\textsc{Concat}([y^s_{\mathcal{M}}, y^s_{\mathcal{M}}]),  f(\mathbf{z}_{\mathcal{M}}; \mathbf{W}_{buf}))$
            \STATE $\alpha \xleftarrow{} \textsc{Sigmoid}(\alpha)$
            \STATE $\mathcal{L}_{DC} \xleftarrow{} \mathcal{L}_{CE_{str}^{\mathcal{D}}} + \alpha \mathcal{L}_{CE_{str}^{\mathcal{M}}} + (1-\alpha) \mathcal{L}_{CE_{buf}^{\mathcal{M}}} $  // Eq. (\ref{eq:dc})
            \STATE $\mathcal{L}_{PAL} \xleftarrow{} \textsc{PAL}(\mathbf{z}_{\mathcal{M}}, \textsc{Concat}([y^s_{\mathcal{M}}, y^s_{\mathcal{M}}]), \mathbf{W}_{str})$ // Eq. (\ref{eq:proxy_anchor_loss})
            \STATE $\mathcal{L}_{Align} \xleftarrow{} \textsc{PAF}
            (\mathbf{W}_{str}, \mathbf{W}_{buf}, \textsc{Concat}([y^{s-1}_{\mathcal{M}}, y^{s-1}_{\mathcal{M}}]))$ // Eq. (\ref{eq:Align})
            \STATE $\Theta \xleftarrow{} \Theta + \lambda \nabla_{\Theta} \mathcal{L}_{BISON}$  // Eq. (\ref{eq:totalloss})
            \STATE $\mathcal{M} \xleftarrow{} \textsc{ReservoirUpdate}(\mathcal{M}, (\mathbf{x}^s_{\mathcal{D}_t},y^s_{\mathcal{D}_t}))$

            \ENDFOR
            
            \ENDFOR
        \STATE Store the label $y^s_{\mathcal{M}}$ 
        \STATE /* Inference Phase: */ 
        \FOR{$l \in C_{1:t}$}
            \STATE $n_l \xleftarrow{}$ number of buffer samples of class $l$
            \STATE $\boldsymbol{\mu}_{l}=\frac{1}{n_{l}} \sum\limits_{(\mathbf{x}, y) \in \mathcal{M}} h\left(\mathbf{x}; \Phi \right) \cdot \mathbb I \left\{{y}=l\right\}$
        \ENDFOR
    \STATE Given a test sample $\mathbf{x}$, $\hat{y} = \underset{l \in C_{1:t}}{\arg \min } \Vert h_{\Phi}(\mathbf{x}) - \boldsymbol{\mu}_l \Vert$ // make a prediction using the NCM classifier
  \ENDFOR
	\end{algorithmic}
\end{algorithm}

\subsection*{Additional Experimental Results}

\begin{table*}[h!]
\caption{Average forgetting at the end of training on three datasets. `T' and `F' indicate two categories: \textit{training distinction} and \textit{feature enhancement}, respectively.}
\centering
\renewcommand{\arraystretch}{1.0}
\resizebox{0.9\textwidth}{!}{%
\begin{tabular}{lccclccclccc}
\hline
\multirow{2}{*}{Method} & \multicolumn{3}{c}{\textbf{Split CIFAR-100}} &  & \multicolumn{3}{c}{\textbf{Split CIFAR-10}} &  & \multicolumn{3}{c}{\textbf{Split Mini-ImageNet}} \\ \cline{2-4} \cline{6-8} \cline{10-12} 
 & $\mathcal{M}$ = 1k & $\mathcal{M}$ = 2k & $\mathcal{M}$ = 5k &  & $\mathcal{M}$ = 0.2k & $\mathcal{M}$ = 0.5k & $\mathcal{M}$ = 1k &  & $\mathcal{M}$ = 1k & $\mathcal{M}$ = 2k & $\mathcal{M}$ = 5k \\ \cline{1-4} \cline{6-8} \cline{10-12} 
FINE-TUNE & \multicolumn{3}{c}{41.4 $_{\pm1.3}$} &  & \multicolumn{3}{c}{68.6 $_{\pm1.7}$} &  & \multicolumn{3}{c}{35.4 $_{\pm0.9}$} \\ \hline
ER & 35.8 $_{\pm1.0}$ & 33.2 $_{\pm1.0}$ & 33.1 $_{\pm1.3}$ &  & 42.9 $_{\pm5.0}$ & 40.0 $_{\pm5.3}$ & 34.2 $_{\pm7.2}$ &  & 32.3 $_{\pm1.4}$ & 30.7 $_{\pm0.9}$ & 31.4 $_{\pm1.2}$ \\
GSS (T) & 34.1 $_{\pm0.9}$ & 32.5 $_{\pm1.7}$ & 32.7 $_{\pm1.0}$ &  & 62.2 $_{\pm1.7}$ & 56.4 $_{\pm3.7}$ & 45.2 $_{\pm5.7}$ &  & 33.2 $_{\pm1.0}$ & 32.8 $_{\pm1.3}$ & 31.4 $_{\pm1.3}$ \\
MIR (T) & 33.9 $_{\pm1.2}$ & 29.9 $_{\pm1.5}$ & 29.4 $_{\pm1.1}$ &  & 40.0 $_{\pm3.9}$ & 31.5 $_{\pm3.4}$ & 23.6 $_{\pm4.3}$ &  & 29.2 $_{\pm1.4}$ & 26.6 $_{\pm1.6}$ & 24.6 $_{\pm1.7}$ \\
A-GEM (T) & 43.0 $_{\pm1.5}$ & 42.1 $_{\pm1.1}$ & 42.8 $_{\pm1.5}$ &  & 67.6 $_{\pm3.4}$ & 65.5 $_{\pm4.2}$ & 67.6 $_{\pm3.8}$ &  & 37.2 $_{\pm0.7}$ & 38.3 $_{\pm1.0}$ & 37.6 $_{\pm0.8}$ \\
Gdumb (T) & 15.4 $_{\pm0.6}$ & 17.2 $_{\pm1.1}$ & 18.0 $_{\pm1.0}$ &  & 23.4 $_{\pm4.0}$ & 21.6 $_{\pm2.6}$ & 19.9 $_{\pm4.1}$ &  & 13.4 $_{\pm0.6}$ & 15.9 $_{\pm0.7}$ & 18.0 $_{\pm1.1}$ \\
SCR (F) & 9.8 $_{\pm0.8}$ & 9.8 $_{\pm0.9}$ & 9.4 $_{\pm0.7}$ &  & 36.8 $_{\pm2.0}$ & 24.5 $_{\pm2.1}$ & 20.6 $_{\pm2.1}$ &  & 8.1 $_{\pm0.6}$ & 7.9 $_{\pm0.7}$ & 7.3 $_{\pm0.6}$ \\
ASER (T) & 37.1 $_{\pm1.6}$ & 33.4 $_{\pm1.4}$ & 27.9 $_{\pm1.4}$ &  & 57.5 $_{\pm2.9}$ & 50.6 $_{\pm4.3}$ & 41.4 $_{\pm4.1}$ &  & 34.5 $_{\pm1.9}$ & 31.2 $_{\pm1.2}$ & 26.4 $_{\pm1.4}$ \\
SS-IL (T) & 10.8 $_{\pm1.3}$ & 11.5 $_{\pm0.9}$ & 10.7 $_{\pm1.4}$ &  & 12.8 $_{\pm3.7}$ & 10.8 $_{\pm2.0}$ & 11.7 $_{\pm2.0}$ &  & 10.9 $_{\pm0.9}$ & 9.7 $_{\pm1.5}$ & 10.3 $_{\pm1.2}$ \\
ER-DVC (T) & 30.8 $_{\pm1.2}$ & 28.0 $_{\pm0.8}$ & 26.0 $_{\pm1.3}$ &  & 32.4 $_{\pm4.6}$ & 28.8 $_{\pm4.6}$ & 19.6 $_{\pm4.6}$ &  & 27.9 $_{\pm1.9}$ & 26.9 $_{\pm0.9}$ & 25.6 $_{\pm1.2}$ \\
ER-ACE (T) & 11.5 $_{\pm1.0}$ & 8.9 $_{\pm0.8}$ & 8.2 $_{\pm0.8}$ &  & 22.4 $_{\pm2.3}$ & 16.6 $_{\pm1.9}$ & 12.8 $_{\pm3.0}$ &  & 11.0 $_{\pm1.6}$ & 8.4 $_{\pm1.5}$ & 7.0 $_{\pm1.6}$ \\
OCM (T, F) & 3.7 $_{\pm0.4}$ & 3.0 $_{\pm0.4}$ & 2.4 $_{\pm0.6}$ &  & 8.2 $_{\pm1.6}$ & 7.6 $_{\pm2.2}$ & 5.0 $_{\pm1.2}$ &  & 3.3 $_{\pm0.5}$ & 2.5 $_{\pm0.4}$ & 2.3 $_{\pm0.6}$ \\
PCR (F) & 15.0 $_{\pm0.9}$ & 11.3 $_{\pm1.7}$ & 8.3 $_{\pm1.2}$ &  & 17.4 $_{\pm5.3}$ & 12.9 $_{\pm2.3}$ & 10.1 $_{\pm2.2}$ &  & 14.5 $_{\pm1.4}$ & 11.3 $_{\pm0.9}$ & 8.9 $_{\pm1.0}$ \\ 
LODE (T) & 15.4 $_{\pm1.4}$ & 13.3 $_{\pm1.0}$ & 11.9 $_{\pm1.6}$ &  & 14.8 $_{\pm1.7}$ & 13.2 $_{\pm3.4}$ & 11.1 $_{\pm2.7}$ &  & 14.8 $_{\pm1.1}$ & 11.8 $_{\pm1.4}$ & 9.8 $_{\pm0.6}$ \\
\hline
\rowcolor{black!10} \textbf{BISON (ours)} & 14.9 $_{\pm1.1}$ & 9.4 $_{\pm0.8}$ & 6.9 $_{\pm0.5}$ &  & 18.8 $_{\pm1.9}$ & 15.4 $_{\pm1.0}$ & 11.5 $_{\pm0.9}$ &  & 10.0 $_{\pm0.6}$ & 7.1 $_{\pm0.5}$ & 5.6 $_{\pm0.5}$ \\
\hline
\end{tabular}%
}
\label{app:AF}
\end{table*}

\begin{table*}[h!]
\caption{\textbf{Average intransigence at the end of training on three datasets.} `T' and `F' indicate two categories: \textit{training distinction} and \textit{feature enhancement}, respectively.}
\centering
\renewcommand{\arraystretch}{1.0}
\resizebox{0.9\textwidth}{!}{%
\begin{tabular}{lccclccclccc}
\hline
\multirow{2}{*}{Method} & \multicolumn{3}{c}{\textbf{Split CIFAR-100}} &  & \multicolumn{3}{c}{\textbf{Split CIFAR-10}} &  & \multicolumn{3}{c}{\textbf{Split Mini-ImageNet}} \\ \cline{2-4} \cline{6-8} \cline{10-12} 
 & $\mathcal{M}$ = 1k & $\mathcal{M}$ = 2k & $\mathcal{M}$ = 5k &  & $\mathcal{M}$ = 0.2k & $\mathcal{M}$ = 0.5k & $\mathcal{M}$ = 1k &  & $\mathcal{M}$ = 1k & $\mathcal{M}$ = 2k & $\mathcal{M}$ = 5k \\ \cline{1-4} \cline{6-8} \cline{10-12} 
FINE-TUNE & \multicolumn{3}{c}{0} &  & \multicolumn{3}{c}{0} &  & \multicolumn{3}{c}{0} \\ \hline
ER & -5.8 $_{\pm3.3}$ & -6.4 $_{\pm4.4}$ & -7.0 $_{\pm3.2}$ &  & 4.9 $_{\pm5.1}$ & 3.2 $_{\pm1.6}$ & 4.9 $_{\pm5.0}$ &  & -6.5 $_{\pm2.5}$ & -6.8 $_{\pm3.0}$ & -5.7 $_{\pm3.0}$ \\
GSS (T) & -4.2 $_{\pm5.0}$ & -4.7 $_{\pm4.9}$ & -4.4 $_{\pm5.1}$ &  & -1.3 $_{\pm3.0}$ & -0.6 $_{\pm1.7}$ & 2.1 $_{\pm3.2}$ &  & -6.2 $_{\pm3.9}$ & -7.4 $_{\pm3.9}$ & -6.0 $_{\pm4.7}$ \\
MIR (T) & -5.3 $_{\pm6.2}$ & -3.6 $_{\pm6.8}$ & -3.0 $_{\pm5.0}$ &  & 8.9 $_{\pm6.2}$ & 10.8 $_{\pm3.4}$ & 18.3 $_{\pm7.5}$ &  & -4.2 $_{\pm7.6}$ & -2.4 $_{\pm6.2}$ & -0.9 $_{\pm7.0}$ \\
A-GEM (T) & -1.8 $_{\pm2.2}$ & -0.6 $_{\pm2.0}$ & -1.9 $_{\pm2.6}$ &  & 0.9 $_{\pm1.6}$ & 3.4 $_{\pm3.8}$ & 0.9 $_{\pm2.8}$ &  & -1.7 $_{\pm2.3}$ & -2.5 $_{\pm2.6}$ & -2.5 $_{\pm2.6}$ \\
Gdumb (T) & 21.0 $_{\pm19.5}$ & 13.2 $_{\pm20.4}$ & -0.4 $_{\pm19.1}$ &  & 36.2 $_{\pm22.4}$ & 29.1 $_{\pm21.2}$ & 22.8 $_{\pm17.2}$ &  & 18.1 $_{\pm17.2}$ & 8.9 $_{\pm19.2}$ & -4.8 $_{\pm19.5}$ \\
SCR (F) & 23.2 $_{\pm11.9}$ & 21.9 $_{\pm10.8}$ & 21.3 $_{\pm11.3}$ &  & 3.1 $_{\pm7.2}$ & 6.7 $_{\pm5.4}$ & 7.7 $_{\pm8.0}$ &  & 19.0 $_{\pm9.2}$ & 17.6 $_{\pm9.3}$ & 16.9 $_{\pm8.7}$ \\
ASER (T) & -9.8 $_{\pm4.3}$ & -8.8 $_{\pm3.0}$ & -6.9 $_{\pm4.3}$ &  & -1.9 $_{\pm2.0}$ & -0.6 $_{\pm3.8}$ & 0.1 $_{\pm4.8}$ &  & -9.0 $_{\pm3.6}$ & -7.8 $_{\pm3.5}$ & -6.4 $_{\pm4.2}$ \\
SS-IL (T) & 15.9 $_{\pm11.6}$ & 12.8 $_{\pm9.5}$ & 14.3 $_{\pm8.4}$ &  & 44.4 $_{\pm29.7}$ & 45.9 $_{\pm23.4}$ & 42.2 $_{\pm29.5}$ &  & 10.6 $_{\pm8.3}$ & 9.8 $_{\pm7.2}$ & 9.8 $_{\pm7.2}$ \\
ER-DVC (T) & -3.7 $_{\pm4.1}$ & -3.6 $_{\pm3.5}$ & -3.4 $_{\pm3.2}$ &  & 8.2 $_{\pm7.7}$ & 12.0 $_{\pm5.2}$ & 15.2 $_{\pm6.3}$ &  & -4.9 $_{\pm4.2}$ & -4.5 $_{\pm2.8}$ & -4.4 $_{\pm3.5}$ \\
ER-ACE (T) & 12.4 $_{\pm10.5}$ & 12.7 $_{\pm11}$ & 11.7 $_{\pm11.2}$ &  & 16.2 $_{\pm15.3}$ & 17.1 $_{\pm11.7}$ & 17.2 $_{\pm12.3}$ &  & 9.3 $_{\pm8.6}$ & 9.4 $_{\pm8.3}$ & 9.2 $_{\pm8.5}$ \\
OCM (T, F) & 38.7 $_{\pm16.1}$ & 38.4 $_{\pm17.6}$ & 38.9 $_{\pm16.6}$ &  & 39.7 $_{\pm32.0}$ & 35.2 $_{\pm33.4}$ & 33.0 $_{\pm30.3}$ &  & 30.0 $_{\pm15.0}$ & 30.6 $_{\pm14.8}$ & 30.2 $_{\pm14.7}$ \\
PCR (F) & 7.3 $_{\pm4.4}$ & 8.6 $_{\pm5.0}$ & 9.5 $_{\pm6.6}$ &  & 21.9 $_{\pm4.4}$ & 25.0 $_{\pm7.2}$ & 25.3 $_{\pm7.3}$ &  & 1.9 $_{\pm5.1}$ & 2.5 $_{\pm6.2}$ & 4.4 $_{\pm7.4}$ \\ 
LODE (T) & 6.7 $_{\pm9.5}$ & 6.7 $_{\pm8.3}$ & 5.7 $_{\pm8.9}$ &  & 21.7 $_{\pm16.4}$ & 16.4 $_{\pm12.8}$ & 18.7 $_{\pm14.3}$ &  & 3.1 $_{\pm8.3}$ & 3.2 $_{\pm7.2}$ & 2.6 $_{\pm8.1}$ \\
\hline
\rowcolor{black!10} \textbf{BISON (ours)} & 5.3 $_{\pm9.7}$ & 7.2 $_{\pm9.5}$ & 6.8 $_{\pm11.4}$ &  & 15.0 $_{\pm13.8}$ & 12.8 $_{\pm10.2}$ & 14.2 $_{\pm14.9}$ &  & 5.9 $_{\pm8.7}$ & 6.2 $_{\pm10.0}$ & 5.8 $_{\pm9.5}$ \\
\hline
\end{tabular}%
}
\label{app:AI}
\end{table*}

\smalltitle{Average Forgetting and Average Intransigence}
In addition to the average accuracy shown in Table \ref{tab:baselines}, we also report the corresponding average forgetting (AF) in Table \ref{app:AF} and the corresponding average intransigence (AI) in Table \ref{app:AI} at the end of training, where all measurements are obtained based on 10 independent runs of experiments. As parts of these results (one buffer for each dataset) are also plotted in Figure \ref{fig:exp_balance}, our BISON method shows the best balance between AF and AI, even though it does not always take the best for both AF and AI at the same time.

\begin{figure*}[t]
	\centering
 \includegraphics[width=0.65\textwidth]{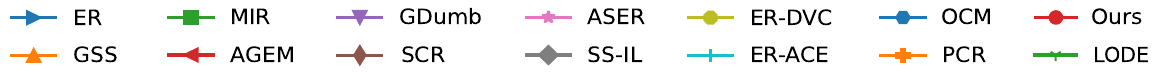} \\
    \subfigure[\label{fig:exp_curve:a}
    {Split CIFAR-100 ($\mathcal{M}$ = 1k)} ]{\hspace{0mm}\includegraphics[width=0.31\textwidth]{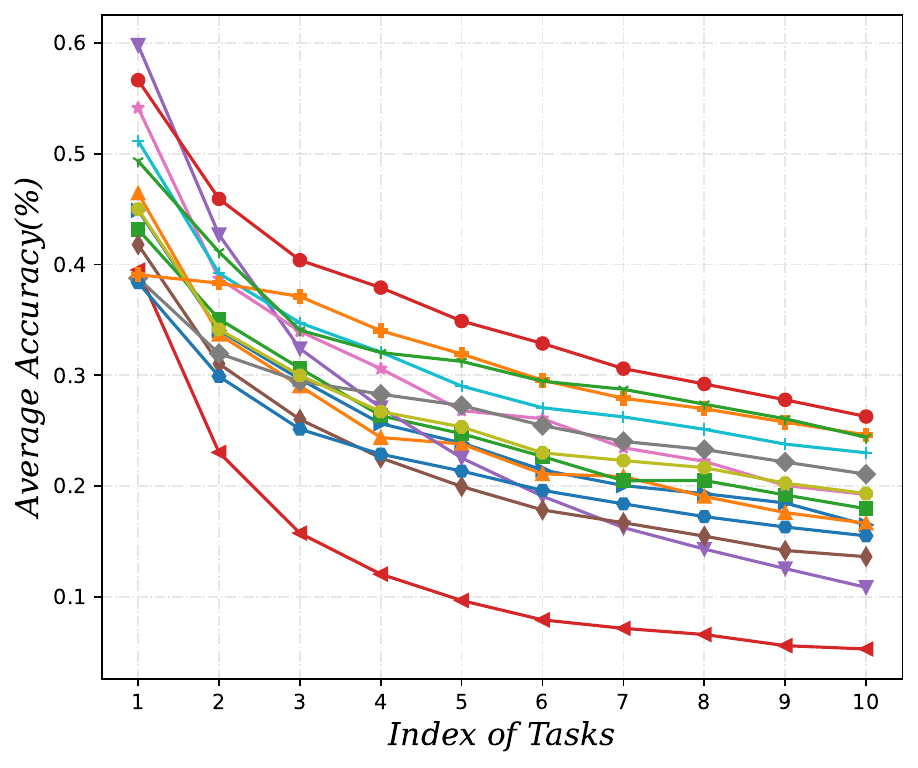}}
    \subfigure[\label{fig:exp_curve:b}{Split CIFAR-10 ($\mathcal{M}$ = 0.5k)} ]{\hspace{0mm}\includegraphics[width=0.31\textwidth]{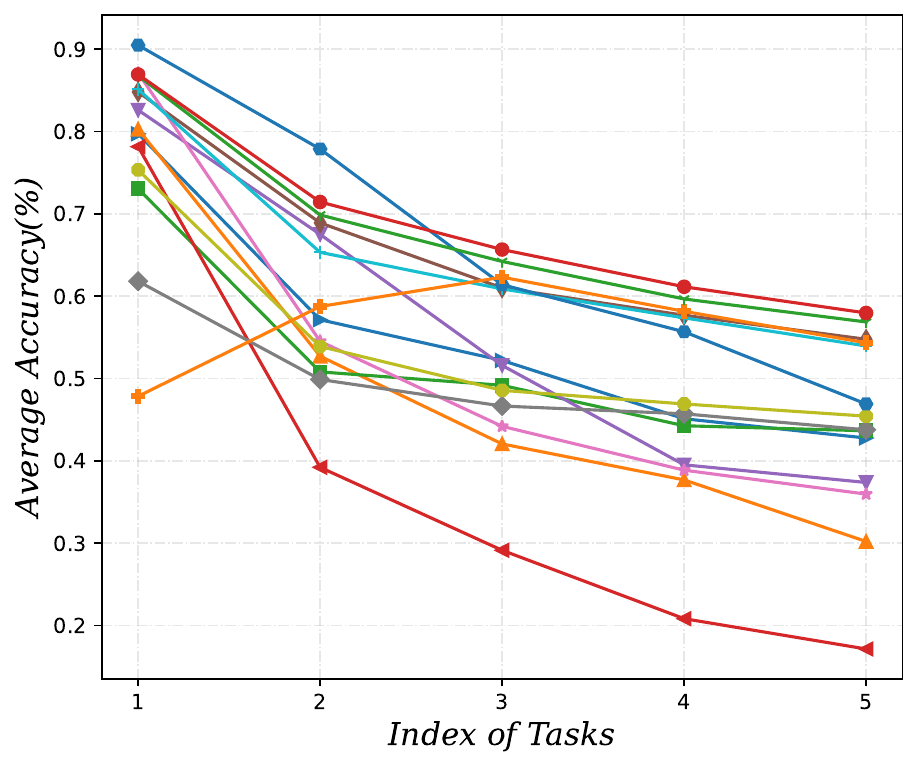}}
    \subfigure[\label{fig:exp_curve:c}{Split Mini-ImageNet ($\mathcal{M}$ = 1k)} ]{\hspace{0mm}\includegraphics[width=0.31\textwidth]{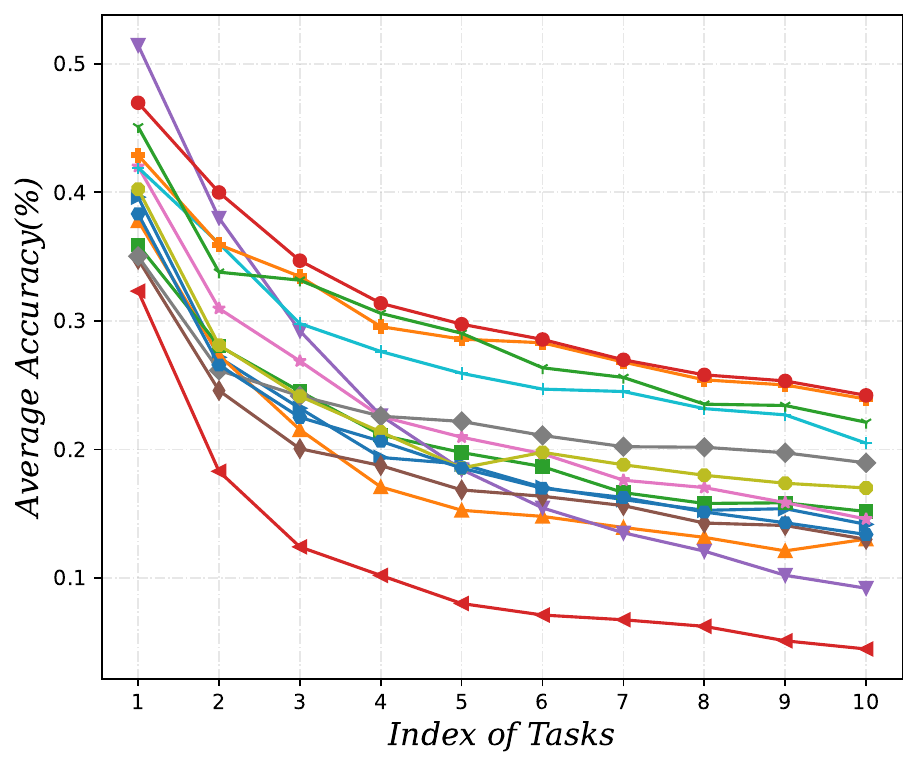}}
    \vspace{-2mm}
    \caption{\textbf{Average accuracy over the incremental training steps.}}
    \label{fig:exp_curve}
\end{figure*}

\smalltitle{Incremental Performance}
Figure \ref{fig:exp_curve} shows how the performance of each method changes over the incremental training steps, measured by average accuracy after learning each task. Our BISON method almost always maintains the highest performance throughout the entire training process, except for the first few tasks. Although Gdumb \cite{Gdumb/PrabhuTD20} starts with the best performance in Split CIFAR-100 and Split Mini-ImageNet, its performance declines sharply after a few subsequent tasks due to both low plasticity and stability, as reported in Figure \ref{fig:exp_balance}. Similarly, OCM, which leads for the first two tasks in Split CIFAR-10, also suffers from a notable performance drop, primarily due to its limited ability of knowledge acquisition, reflected in its low plasticity in Figure \ref{fig:exp_balance}. Meanwhile, PCR demonstrates competitive performance overall in Table \ref{tab:baselines} and Figure \ref{fig:exp_balance}, but struggles with initial tasks, exhibiting considerably lower performance compared to other methods. In contrast, BISON starts effectively enough and continues to outperform other methods in subsequent tasks.

\begin{figure*}[t]
\centering
\includegraphics[height=50mm]{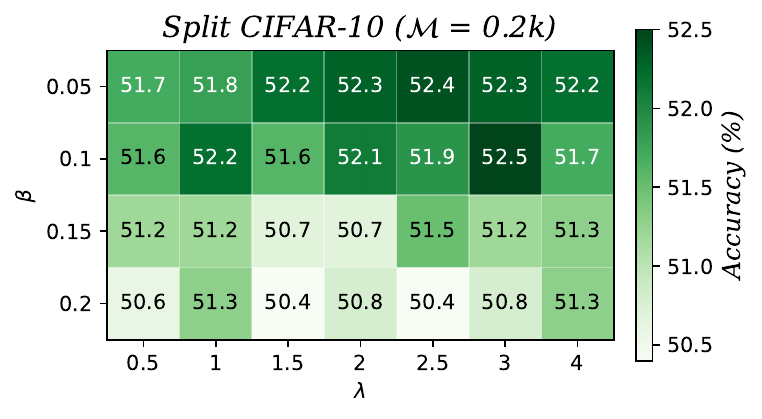} \hspace{3mm}
\includegraphics[height=50mm]{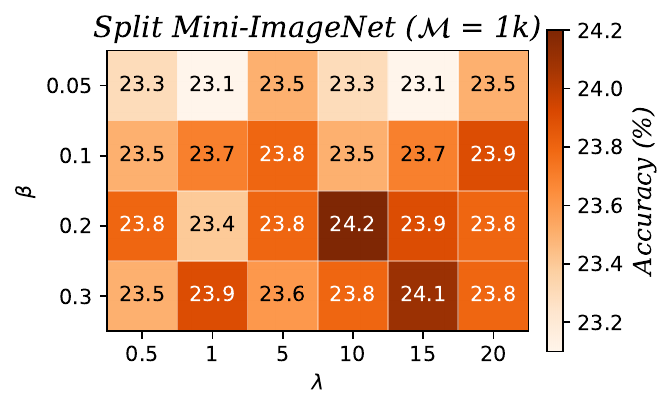} \hspace{3mm}
\vspace{-4mm}
\caption{Impact of Hyperparameter $\lambda$ and $\beta$ on Split CIFAR-10 ($\mathcal{M}$ = 0.2k) and Split Mini-ImageNet($\mathcal{M}$ = 1k).}
\label{fig:hyperparameter}
\end{figure*}

\smalltitle{Hyperparameter Sensitivity}
We conduct a hyperparameter sensitivity analysis on Split CIFAR-10 ($\mathcal{M}$ = 0.2k) and Split Mini-ImageNet ($\mathcal{M}$ = 1k), as shown in Figure \ref{fig:hyperparameter}. All results are averaged over 10 independent runs. For the smaller dataset, Split CIFAR-10, we examine two hyperparameters, $\beta$ and $\lambda$, ranging over $[0.05, 0.1, 0.15, 0.2]$ and $[0.5, 1, 1.5, 2, 2.5, 3, 4]$, respectively. A clear performance plateau appears around $\lambda \in [2, 3]$ with $\beta \in [0.05, 0.1]$. Along this plateau, varying $\lambda$ causes only minor fluctuations in average accuracy, whereas increasing $\beta$ from 0.1 to 0.2–0.3 consistently degrades performance. The redesigned Proxy-Anchor loss $\mathcal{L}{PAL}$ is computed between the buffer features $\mathbf{z}{\mathcal{M}}$ and the proxies in $\mathbf{W}_{str}$, and its gradients propagate into both the shared backbone and the stream classifier. Consequently, the stream classifier is indirectly influenced by exemplars of old classes. As $\beta$ increases, it enhances the discriminative ability of the stream classifier, but an excessively large $\beta$ disturbs the stability and overall balance, ultimately degrading performance. Overall, Split CIFAR-10 is more sensitive to $\beta$ than to $\lambda$. On Split Mini-ImageNet, due to the limited number of samples in each buffer batch and the larger number of proxies in the stream classifier, stronger values of $\beta$ and $\lambda$ are required to facilitate effective knowledge interaction. Split Mini-ImageNet exhibits a plateau when $\lambda \approx 10$–$15$ with moderate $\beta \in [0.2, 0.3]$, where performance becomes relatively more sensitive to $\lambda$. Varying $\beta$ within 0.2–0.3 changes accuracy only slightly. We also explore the balance on these two datasets and find that the best balance is achieved near the same regions as the best overall performance ($52.5\pm0.7\%$ at $(\lambda,\beta)=(3, 0.1)$ and $24.2\pm0.4\%$ at $(10, 0.2)$). In conclusion, BISON shows low sensitivity to hyperparameters, as evidenced by the wide performance plateaus. For Split CIFAR-100, which contains the same number of classes, we directly adopt the best settings from Split Mini-ImageNet.

\smalltitle{Comparison Under Another Widespread Setting for Feature Enhancement Methods}
OCM \cite{OCM/GuoLZ22} and OnPro \cite{OnPro/0001YHZS23} adopt several proprietary data augmentation techniques such as local and global rotations, and they rely on stronger backbones and larger buffer batches. Moreover, OnPro only supports learning in fixed label order and SCR \cite{SCR/MaiLKS21} is also sensitive to the size of the buffer batch. 
To ensure fairness, we adapt our BISON to their settings and validate the performance on Split CIFAR-100 under varying memory buffer sizes. We employ ResNet-18 \cite{ResNet/HeZRS16} (with 64 filters) and set the size of each buffer batch to 64. The training tasks and label order are fixed, and  results are averaged over 15 independently runs. 
The results of OCM are taken directly from their original paper, while OnPro only reports results for $\mathcal{M}$ = 1k and 2k. Therefore, we mark these results with $\dag$ and reproduce their method with $\mathcal{M}$ = 5k for comparison. 
As shown in Table \ref{tab:Comparison of FE}, our method BISON consistently achieves the best performance across varying memory buffer sizes, outperforming other feature enhancement methods.

\begin{table}[t]
\caption{Comparison under another widespread setting for feature enhancement methods on Split CIFAR-100.}
\centering
\resizebox{0.4\textwidth}{!}{%
\begin{tabular}{llll}
\hline
\multirow{2}{*}{Methods} & \multicolumn{3}{c}{Split CIFAR-100} \\ \cline{2-4} 
 & \multicolumn{1}{c}{$\mathcal{M}$ = 1000} & \multicolumn{1}{c}{$\mathcal{M}$ = 2000} & \multicolumn{1}{c}{$\mathcal{M}$ = 5000} \\ \hline
SCR & $26.3\pm0.6$ & $31.7\pm0.4$ & $35.3\pm0.4$ \\
OCM & $28.1\pm0.3$ & $35.0\pm0.4$ & $42.4\pm0.5$ \\
OnPro$\dag$ & $30.0\pm0.4$ & $35.9\pm0.6$ & $40.9\pm0.2$ \\
PCR & $29.8\pm0.7$ & $36.3\pm0.6$ & $43.5\pm0.5$ \\
\textbf{BISON(ours)} & \bm{$33.5\pm0.9$} & \bm{$40.4\pm0.8$} & \bm{$45.5\pm1.3$} \\ \hline
\end{tabular}%
}
\label{tab:Comparison of FE}
\end{table}

\begin{figure}[t!]
\centering
\includegraphics[width=0.5\textwidth]{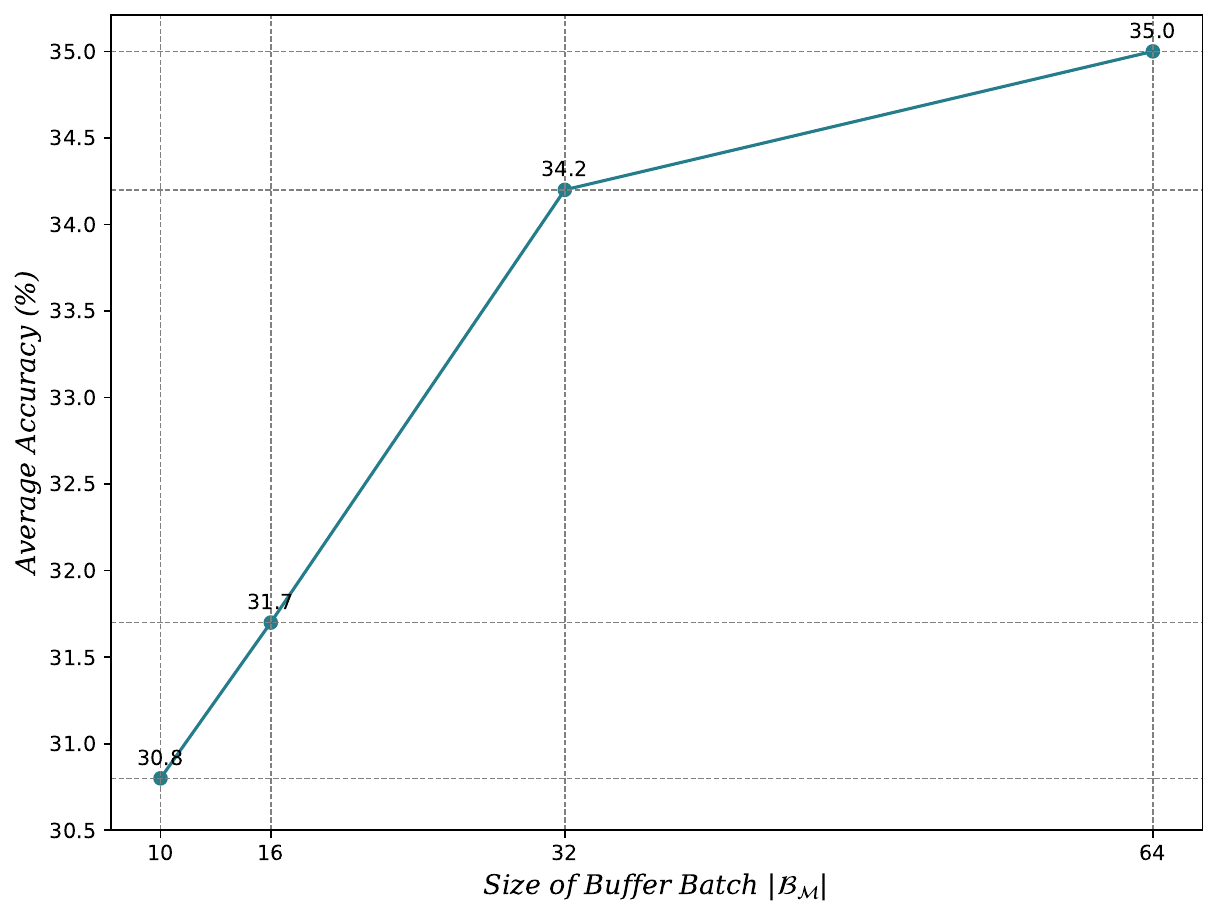}
\vspace{1mm}
\caption{\textbf{Average accuracy of our BISON on Split CIFAR-100 ($\mathcal{M}$ = 2k) with various sizes of buffer batch.}}
\label{fig:app_buffersize}
\vspace{1mm}
\end{figure}

\begin{table}[t]
\caption{\textbf{Analysis of computational efficiency on Split CIFAR-10 ($\mathcal{M}$ = 1k) and Split CIFAR-100 ($\mathcal{M}$ = 5k) datasets.}}
\centering
\resizebox{0.70\textwidth}{!}{%
\begin{tabular}{lcccc}
\hline
\multirow{2}{*}{Method} & \multicolumn{2}{c}{Split CIFAR-10 ($\mathcal{M}$ = 1k)} & \multicolumn{2}{c}{Split CIFAR-100 ($\mathcal{M}$ = 5k)} \\ \cline{2-5} 
 & Num of Params. & GPU Usage(Mb) & Num of Params. & GPU Usage(Mb) \\ \hline
ER & 1094750 & 1787 & 1109240 & 1837 \\
BISON & 1096343 & 1805 & 1125143 & 1853 \\ \hline
\end{tabular}%
}
\label{tab:comp_efficiency}
\end{table}

\smalltitle{Impact of Size of Buffer Batch $|\mathcal{B}_\mathcal{M}|$ }
In our experiments, buffer batch size $|\mathcal{B}_\mathcal{M}|$ is set to $10$, which is the same as the size of stream batch. As shown in Figure \ref{fig:app_buffersize}, we examine the impact of buffer batch size on classification performance. As the buffer batch size becomes larger, the performance gradually improves. However, when it is larger than $32$, the improvement is limited.
Remarkably, when the model is trained in an online setting, both efficiency and performance need to be taken into consideration, and hence it is not always feasible to set this value too large.

\smalltitle{Computational Efficiency}
As presented in Figure \ref{tab:comp_efficiency}, We investigate changes in parameters and GPU usage of our BISON method. Compared to baseline ER, BISON increases model size by only 0.1\% (Split CIFAR-10) and 1.4\% (Split CIFAR-100), with GPU usage increasing less than 1\%. Since additional classification head is small relative to the backbone, the impact on memory and inference latency is negligible. Overall, our BISON method is computationally efficient and feasible.

\subsection*{Full Implementation Details}
Stochastic gradient descent (SDG) optimizer is used for training the parameters from scratch with an initial learning rate $\lambda=$ 0.1 for all experiments except for OCM \cite{OCM/GuoLZ22}. Note that the classes are shuffled in each task for every run. As mentioned in the main paper, for our BISON method, we set $\gamma=32$ and $\delta=0.1$ in Eq. (\ref{eq:totalloss}) for all datasets. 
For Split CIFAR-100 and Split Mini-ImageNet, $\beta$ and $\lambda$ in Eq. (\ref{eq:totalloss}) are set to $0.2$ and $10.0$, respectively, while they are set to $0.1$ and $3.0$, respectively, for Split CIFAR-10.

\smalltitle{Data Augmentation}
We also follow existing works such as \cite{PCR/LinZFLY23} for data augmentation. More specifically, the scale in resized-crop is set to $(0.2, 1)$ for each datasets. In color-jitter, the (brightness, contrast, saturation, hue) is set to $(0.4, 0.4, 0.4, 0.1)$ with probability $0.8$. The probability of grey-scale is $0.2$. These data augmentation techniques are applied to both stream and buffer samples for all methods with the exception of OCM, as OCM takes advantage of their own data augmentation techniques.

\smalltitle{Reproduction}
For ensuring reproducibility, the random seed for all experiments is fixed to be $0$, and we use default experimental settings like hyperparameter values for reproducing the results of other compared methods. Unless otherwise notified, we use random sampling to retrieve 10 buffer samples, and employ reservoir updating for buffer management, which are generally adopted by existing replay-based methods. For GSS \cite{GSS/AljundiLGB19}, we randomly sample $10$ buffer batches to estimate the maximum cosine similarity scores. For MIR \cite{MIR/AljundiBTCCLP19}, the number of sub-sample in their memory retrieval strategy is $50$. For Gdumb \cite{Gdumb/PrabhuTD20}, the gradient clip is set to $10$ and we train the memory buffer for $30$ epochs. The data augmentation is applied on their memory training. Memory updating strategy is their greedy balancing updating. For SCR \cite{SCR/MaiLKS21}, the temperature $\tau$ in supervised contrastive loss is $0.07$. The projection head is multi-layer perceptron with feature dimensions of $128$. For ASER \cite{ASER/ShimMJSKJ21}, the memory updating and retrieval strategy is replaced by their own proposed SV-based strategy. For ER-DVC \cite{DVC/Gu0WD22}, the number of sub-sample in their MGI retrieval strategy is $50$, the coefficient of $\lambda_3 = 2$ for Split CIFAR-10, while $\lambda_3 = 4$ for Split CIFAR-100 and Split Mini-ImageNet. We follow the setting $\lambda_1=\lambda_2=1$ for all datasets. For OCM \cite{OCM/GuoLZ22}, we use the reduced ResNet-18, and Adam optimizer is utilized with $0.001$ learning rate, $betas = (0.9, 0.99)$ and $0.0001$ weight decay as in their paper. For data augmentation in OCM, resized-crop is set to $(0.3,0.1)$, and the probability of grey-scale is $0.25$. The horizontal-flip, the local rotation and global rotation are also employed. For Split CIFAR-10 and CIFAR-100 in OCM, we follow their hyperparameter values. For Split Mini-ImageNet, we use the same settings as those in Split CIFAR-100. Their max $N_b$ is set to 10. For PCR \cite{PCR/LinZFLY23}, we use the temperature $\tau = 0.07$ in supervised contrastive loss. For LODE \cite{LODE/LiangL23}, we take the best hyperparameter values from their paper, where it is equipped with DER++\cite{DERPP/BuzzegaBPAC20} and $\alpha = 0.1$, $\beta = 1.0$ and $\rho = 0.1$ for Split CIFAR-10. As for Split CIFAR-100, $\alpha = 0.2$, $\beta = 0.2$ and $\rho = 0.5$ and we keep the same setting for Split Mini-ImageNet. For all experiments of LODE, the learning rate for optimizing is $0.03$ and the $\beta_1 = C = 1.0$ as default.


\end{document}